\documentclass[mnsc,nonblindrev]{informs4} 

\OneAndAHalfSpacedXI

\definecolor{darkgreen}{rgb}{0,0.35,0}

\usepackage{subcaption}
\usepackage{mathrsfs}
\usepackage{amsmath,bm}
\usepackage{graphicx}

\usepackage{booktabs}
\usepackage{url}
\definecolor{DarkBlue}{rgb}{0,0.08,0.45}
\usepackage[backref = false, bookmarks, colorlinks = true, plainpages = false, citecolor = DarkBlue , urlcolor = DarkBlue, filecolor = DarkBlue, linkcolor = DarkBlue]{hyperref}
\usepackage{natbib}
\usepackage{multibib}
\newcites{EC}{References} 

\usepackage{newunicodechar}
\usepackage[utf8]{inputenc}
\usepackage{algorithm}

\newcommand{\rmnum}[1]{\romannumeral #1}
\usepackage{natbib}
 \bibpunct[, ]{(}{)}{,}{a}{}{,}%
 \def\bibfont{\small}%

 \usepackage{pgfplots}
\usepgfplotslibrary{fillbetween}
\pgfplotsset{compat=1.13}
\usepackage{algpseudocode}
\TheoremsNumberedThrough     
\ECRepeatTheorems

\EquationsNumberedThrough    

\begin{document}
\newcommand{\abs}[1]{\left|  #1 \right| }
\newcommand{\brak}[1]{\left(#1\right)}    
\newcommand{\crl}[1]{\left\{#1\right\}}   
\newcommand{\edg}[1]{\left[#1\right]}     
\newcommand{\norm}[1]{\|#1\|}
\newcommand{\floor}[1]{\lfloor #1 \rfloor}

\newcommand{\cA}{{\mathcal A}}
\newcommand{\cB}{{\mathcal B}}
\newcommand{\cD}{{\mathcal D}}
\newcommand{\cF}{{\mathcal F}}
\newcommand{\cG}{{\mathcal G}}
\newcommand{\cH}{{\mathcal H}}
\newcommand{\cK}{{\mathcal K}}
\newcommand{\cL}{{\mathcal L}}
\newcommand{\cM}{{\mathcal M}}
\newcommand{\cR}{{\mathcal R}}
\newcommand{\cS}{{\mathcal S}}
\newcommand{\cT}{{\mathcal T}}
\newcommand{\cX}{{\mathcal X}}
\newcommand{\cP}{{\mathcal P}}
\newcommand{\cV}{{\mathcal V}}

\newcommand{\mA}{{\mathbb A}}
\newcommand{\mV}{{\mathbb V}}
\newcommand{\mC}{{\mathbb C}}
\newcommand{\mR}{{\mathbb R}}
\newcommand{\mE}{{\mathbb E}}
\newcommand{\mw}{{\mathbb w}}
\newcommand{\mT}{{\mathbb T}}

\newcommand{\bb}{{\mathbf b}}
\newcommand{\bd}{{\boldsymbol d}}
\newcommand{\by}{{\mathbf y}}
\newcommand{\bI}{{\mathbf I}}
\newcommand{\bp}{{\mathbf p}}
\newcommand{\bc}{{\mathbf c}}
\newcommand{\bg}{{\mathbf g}}
\newcommand{\bl}{{\mathbf l}}
\newcommand{\bbf}{{\mathbf f}}
\newcommand{\bq}{{\mathbf q}}

\newcommand{\bcD}{{\boldsymbol{\mathcal{D}}}}
\newcommand{\bbD}{{\boldsymbol{\mathscr{D}}}}

\newcommand{\bx}{{\boldsymbol x}}
\newcommand{\bA}{{\mathbf A}}
\newcommand{\bB}{{\mathbf B}}
\newcommand{\bC}{{\mathbf C}}
\newcommand{\bD}{{\mathbf D}}
\newcommand{\bG}{{\mathbf G}}
\newcommand{\bL}{{\mathbf L}}
\newcommand{\bS}{{\mathbf S}}
\newcommand{\bQ}{{\boldsymbol Q}}
\newcommand{\bU}{{\mathbf U}}
\newcommand{\bV}{{\boldsymbol V}}
\newcommand{\bK}{{\boldsymbol K}}
\newcommand{\bX}{{\mathbf X}}
\newcommand{\bZ}{{\mathbf Z}}
\newcommand{\bF}{{\mathbf F}}

\newcommand{\bmu}{{\boldsymbol \mu}}
\newcommand{\bomega}{{\boldsymbol \omega}}
\newcommand{\bw}{{\boldsymbol w}}
\newcommand{\bW}{{\boldsymbol W}}
\newcommand{\balpha}{{\boldsymbol \alpha}}
\newcommand{\blambda}{{\boldsymbol \lambda}}
\newcommand{\bxi}{{\boldsymbol \xi}}

\newcommand{\btheta}{\boldsymbol{\theta}}
\newcommand{\bsigma}{\boldsymbol{\sigma}}
\newcommand{\bnu}{\boldsymbol{\nu}}
\newcommand{\bSigma}{\boldsymbol{\Sigma}}
\newcommand{\bgamma}{\boldsymbol{\gamma}}
\newcommand{\bs}{\boldsymbol{s}}
\newcommand{\bz}{\boldsymbol{z}}

\newcommand{\C}{\mathbb{C}}

\newcommand{\D}{\mathbb{D}}
\newcommand{\F}{\mathbb{F}}
\newcommand{\p}{\mathbb{P}}
\newcommand{\Q}{\mathbb{Q}}
\newcommand{\W}{\mathbb{W}}
\newcommand{\R}{\mathbb{R}}
\newcommand{\q}{\mathbb{Q}}

\newcommand{\tr}{{\rm tr}}

\newcommand{\id}{{\mathbbm 1}}

\newcommand{\expect}{\mathbb{E}}



\RUNTITLE{
LLM-Inspired Pretrain-Then-Finetune for Small-Data Large-Scale Optimization
}

\TITLE{LLM-Inspired Pretrain-Then-Finetune \\for Small-Data, Large-Scale Optimization}


\ARTICLEAUTHORS{%
\AUTHOR{Zishi Zhang}
\AFF{Guanghua School of Management, Peking University, Beijing 100871, China, \EMAIL{zishizhang@stu.pku.edu.cn}}
\AUTHOR{Jinhui Han}
\AFF{Guanghua School of Management, Peking University, Beijing 100871, China, \EMAIL{jinhui.han@gsm.pku.edu.cn}}
\AUTHOR{Ming Hu}
\AFF{Rotman School of Management, University of Toronto,  Toronto, Ontario, Canada M5S 3E6, \EMAIL{ming.hu@rotman.utoronto.ca}}
\AUTHOR{Yijie Peng}
\AFF{Guanghua School of Management, Peking University, Beijing 100871, China, \EMAIL{pengyijie@pku.edu.cn}}
} 

\ABSTRACT{%
We consider small-data, large-scale decision problems in which a firm must make many operational decisions simultaneously (e.g., across a large product portfolio) while observing only a few, potentially noisy, data points per instance. Inspired by the success of large language models (LLMs), we propose a pretrain-then-finetune approach built on a designed Transformer model to address this challenge. The model is first pretrained on large-scale, domain-informed synthetic data that encode managerial knowledge and structural features of the decision environment, and is then fine-tuned on real observations. This new pipeline offers two complementary advantages: pretraining injects domain knowledge into the learning process and enables the training of high-capacity models using abundant synthetic data, while finetuning adapts the pretrained model to the operational environment and improves alignment with the true data-generating regime. While we have leveraged the Transformer's state-of-the-art representational capacity, particularly its attention mechanism, to efficiently extract cross-task structure, our approach is not an off-the-shelf application. Instead, it relies on problem-specific architectural design and a tailored training procedure to match the decision setting. Theoretically, we develop the first comprehensive error analysis regarding Transformer learning in relevant contexts, establishing nonasymptotic guarantees that validate the method’s effectiveness. Critically, our analysis reveals how pretraining and fine-tuning jointly determine performance, with the dominant contribution governed by whichever is more favorable. In particular, finetuning exhibits an economies-of-scale effect, whereby transfer learning becomes increasingly effective as the number of instances grows. The results also highlight the important role of domain-guided pretraining, which provides an informative initialization that improves performance, especially when domain knowledge is accurate and the real data are extremely limited. These insights are echoed in numerical experiments and, more broadly, shed light on the effectiveness of Transformer-based LLM training.
}%



\maketitle

%


\section{Introduction}\label{sec:introduction}

Transformer-based models \citep{vaswani2017attention} have achieved remarkable success across a broad range of scientific domains, including natural language processing \citep{wolf2020transformers}, protein structure prediction \citep{jumper2021highly}, and even mathematical discovery \citep{romera2024mathematical}. In particular, large language models (LLMs) stand out as perhaps the most visible and widely adopted instantiation of this model, with prominent examples including the GPT family~\citep{gpt3}, BERT~\citep{bert}, and Deepseek \citep{liu2024deepseek}. From a methodological perspective, the pretrain-then-finetune paradigm built on Transformer architectures has emerged as a central principle behind the success of LLMs. Under this framework, a model is first pretrained via (self-)supervised learning on large-scale raw-text corpora, enabling it to acquire rich, transferable representations. The pretrained model is then finetuned using task-specific data to adapt these representations to downstream objectives.

LLMs are, as the name implies, essentially very large; for example, GPT-3 contains roughly 175 billion parameters, and Deepseek-V3 activates about 37 billion parameters per token during inference. Such a scale is enabled by access to massive text corpora, and the model serves a general-purpose set of objectives, spanning information retrieval and question answering to increasingly sophisticated forms of reasoning. Business decision-making settings differ in several key aspects. The available data are typically far more limited; in many cases, they are also private and highly sensitive. Moreover, the decision objectives are well specified (e.g., minimizing inventory costs or maximizing revenue in pricing) rather than open-ended. These features make a smaller, specialized model both practical and, often, necessary: it can be trained feasibly on limited data while directing modeling capacity toward the decision-relevant structure. Motivated by these considerations and by the success of LLM training pipelines, we study the pretrain-then-finetune approach for small-data, large-scale stochastic optimization. Our method builds on a designed Transformer architecture, pretrained using domain-informed (synthetic) data and subsequently finetuned on scarce real observations. We further establish quantitative theoretical guarantees that characterize its effectiveness for the target optimization problem.

Specifically, we focus on operations management (OM) problems in which firms must manage vast product catalogs, yet most items lie in the \emph{long tail} \citep{anderson2006long} and generate only sparse sales histories. This phenomenon, referred to as a small-data large-scale problem in \cite{mivsic2020data}, poses a central challenge for deploying data-driven methods in relevant OM scenarios. For instance, using data from Tmall, \cite{miao2022context} report that among more than 75,000 products listed between May and July 2018, over 21.6\% attracted fewer than 10 unique visitors per day on average, and more than 14.3\% received at most one visitor per day. A similar pattern is observed in fast fashion: Shein reportedly launches as many as 10,000 new SKUs per day, typically with small initial batches of roughly 100 units, and many products have life cycles of approximately 40 days \citep{kale2023shein}. In such environments, decisions must be made before sufficient sales data have accumulated, even though the decision space is enormous. More broadly, data limitations also arise in moderate-scale problems due to privacy constraints, short histories, or infrequent observations. These unique characteristics limit the applicability of traditional approaches based on direct sample average approximation (SAA) and, typically, render them operationally infeasible.

The pretrain-then-finetune idea, originally developed in LLM training, is well-suited to our target problem, which is formulated as solving $N$ data-driven stochastic optimization problems simultaneously, each governed by a parameterized distribution with unknown parameters. 
It proceeds in two stages.
First, pretraining leverages abundant raw or synthetic data and, as evidenced in LLM applications, can yield strong performance even when task-specific data are limited. In OM contexts, it offers two particular advantages. On the one hand, it makes training feasible in settings where proprietary business data are scarce, thereby enabling the use of expressive modern AI models. On the other hand, it provides a disciplined way to inject domain knowledge into learning. For example, one can construct synthetic data from managerial judgment, stylized models in the literature, or empirical regularities from historical experience; more recently, generative models have further expanded the ability to produce high-quality pretraining data tailored to problem structure (see, e.g., \citealt{hollmann2025accurate}).
Second, fine-tuning adapts the pretrained model to the specific operational environment using available real-world observations. Consistent with the small-data regime, fine-tuning updates only a subset of parameters while keeping others fixed, enabling more precise adaptation while leveraging useful knowledge learned during pretraining. Taken together, the two-stage procedure yields an approach that is both domain-informed and empirically grounded. Operationally, we implement this pipeline using a tailored Transformer architecture that handles variable-length problem instances and infers the latent parameters underlying each instance. These parameter estimates are then mapped to the corresponding optimal decisions. To the best of our knowledge, both the solution approach and the associated problem-solving tool are new to the OM literature.

Although our pretrain-then-finetune approach is inspired by LLM training, it is \emph{not} a direct application of off-the-shelf LLM tools; Section \ref{sec:method} provides the details of our implementation and highlights the key methodological design. Instead, we develop a compact, task-specific Transformer tailored to the problem structure. In our framework, the Transformer primarily serves as a parameter estimator. During pretraining, it is trained in a supervised manner on synthetic datasets with known labels that encode domain knowledge. In finetuning, however, supervision is unavailable because the ground-truth parameters are unobserved. To address this challenge, we construct a mean squared error (MSE)-equivalent objective using a generalized Stein identity, which yields a tractable loss whose minimization aligns with minimizing the parameter estimation error. To further accommodate the small-data regime, we adopt low-rank adaptation (LoRA; \citealt{hu2022lora}) to update only a selected subset of Transformer parameters, thereby reducing data requirements and improving computational efficiency. Specifically, we emphasize that the Transformer is not intended as a generic function approximator that could be readily replaced by standard deep neural networks (DNNs). Its attention mechanism is particularly well suited to our setting because it enables the model to aggregate information across heterogeneous instances and transfer learned structure to new problems---capabilities that are central to the effectiveness of the pretrain–finetune paradigm in our target problem.

What about theoretical guarantees, i.e., can we formally validate efficiency, build confidence for practical deployment, and provide insights into understanding its performance? We provide a comprehensive error analysis, from the estimation error of the pretrained-then-finetuned Transformer to the resulting decision quality, which compares the induced decisions with an oracle benchmark that has prior access to the shared meta-information across tasks. This is, to our knowledge, the first analysis to comprehensively characterize the error mechanisms of Transformer-based models in the learning-theoretic literature. More broadly, despite extensive empirical evidence on how LLMs perform well (see, e.g., \citealt{kaplan2020scaling,brown2020language}), theoretical understanding remains very limited, especially given that these systems are substantially larger and more complex ``black boxes" than standard AI models such as DNNs. To this end, while our analysis does not aim to explain the full behavior of modern LLMs, it provides principled insights into LLMs by focusing on their core architectural primitive, i.e., the Transformer, and by establishing guarantees that link representation learning to end-to-end decision performance.

The theoretical analysis in Section \ref{sec:theory} begins with a general principle that decomposes the estimation error into three different sources: (i) a pretraining domain gap, which captures the mismatch between pretraining knowledge and the operational environment; (ii) a finetuning generalization error, which quantifies how well rules learned from finite samples generalize to new instances; and (iii) an approximation error, which reflects representational limitations of the model class. A key insight is that either the domain gap or the generalization error can be the dominant contributor, highlighting the substantive interaction between pretraining and finetuning. Specifically, the domain gap decreases as the distributional discrepancy between the pretraining data and the true stochastic environment narrows. Importantly, the finetuning generalization error decreases with problem scale, at a rate no slower than $\mathcal{O}(1/\sqrt{N})$. This characterization yields a natural interpretation: when domain knowledge is reliable but real data are scarce, the model’s output is primarily guided by pretraining; as additional observations become available, finetuning increasingly aligns the learned representations with practice. Even in the latter regime, effective pretraining remains valuable by improving the efficiency and stability of finetuning and, consequently, the final decision performance. Two additional implications are worth emphasizing. First, the generalization error scales with the number of tasks $N$, which formally supports the ability of our approach to transfer information across tasks in an automated manner. Second, the approximation error decreases with model capacity, providing a theoretical rationale for the effectiveness of larger architectures when abundant pretraining data are available. Finally, we balance these error components to guide network design and to derive an excess-risk bound for the induced decisions. 
In summary, this new comprehensive analysis provides theoretical guarantees for our approach and contributes to the learning-theoretic understanding of Transformer models in data-driven decision problems.

Numerically, we evaluate the proposed approach on a multi-product newsvendor problem, using it as a testbed to assess performance and to verify the theoretical insights on pretraining–finetuning interaction developed in Section~\ref{sec:theory}.
We find: (1) Synthetic-data pretraining is essential for training large Transformer models when task-specific observations are scarce, and the quality of the pretrained estimator depends critically on how well the decision maker’s domain knowledge aligns with the true environment;
(2) When this knowledge is sufficiently accurate, the pretrained model already performs strongly, so additional finetuning on real data provides limited incremental benefit; and
(3) When domain knowledge is misspecified, finetuning becomes indispensable: by leveraging cross-task transfer, the finetuned model gradually mitigates the bias inherited from pretraining. Moreover, as the problem scale grows, finetuning becomes increasingly effective, and the resulting decisions approach the oracle benchmark. These qualitative patterns remain robust in various data-generating settings.



\section{Literature Review}\label{sec:literature}

This paper contributes to the development of data-driven AI methodologies for operations management, with a particular focus on state-of-the-art Transformer-based techniques inspired by recent advances in LLMs. Our work relates to two primary streams of the literature.

The first stream studies how machine learning and data can be leveraged to support OM decisions. This literature has expanded rapidly in recent years, driven by the growing availability of data and advances in algorithms and computational resources. A central theme is to extract decision-relevant information from historical observations and translate it into prescriptive policies for the underlying operational problem. Two broad regimes are commonly distinguished: ``big-data" and ``small-data" settings. In the big-data regime, one can approximate the system’s stochasticity using the empirical distribution, so the sample size for each decision problem must be sufficiently large to ensure statistical accuracy. For example, \citet{levi2015data} analyze SAA for the data-driven newsvendor problem and establish finite-sample performance guarantees; \citet{lin2022data} extend this line of analysis in a more general framework. Beyond SAA, several papers incorporate additional information structures. \citet{ban2019big} show that contextual information can improve inventory decisions, and \citet{han2023deep} demonstrate that DNNs can further enhance such performance. \citet{chen2023model} learn customer preferences from transaction data to design assortment and pricing decisions for new customers. More broadly, \citet{bertsimas2020predictive}, \citet{chen2025using}, and \citet{ETOvsIEO} study general stochastic optimization problems with predictive components. A complementary set of papers investigates the consequences of misalignment between estimation and optimization: \citet{siegel2021profit,siegel2023data} investigate potential bias when learning objectives are not well aligned with optimization objectives, and \citet{OIC} propose a statistical criterion that can be used to detect and correct such bias in general data-driven optimization frameworks.

In the small-data regime, each problem instance is associated with only a limited number of observations, even though the total data volume aggregated across instances can be substantial \citep{mivsic2020data}. This setting is common and important in practice, particularly in platforms and long-tail retail environments \citep{miao2022context}. The key challenge is that the per-instance sample size is too small to support a reliable empirical approximation of the underlying uncertainty for each individual optimization problem. \citet{gupta2021small} study linear optimization under small-data constraints and propose a Bayes-regularized policy with theoretical guarantees. \citet{gupta2022data} show that effective data pooling via Shrunk-SAA
can improve general performance. \citet{bastani2022meta} examine pricing with sequential product offerings and demonstrate how meta-learning can be used to transfer knowledge across products. In the context of low-sale items on Alibaba, \citet{miao2022context} propose a clustering-based pricing algorithm that pools data within clusters to improve learning and decision quality. \citet{feng2023transfer} leverage structural relationships between data and optimal solutions to facilitate transfer learning across problems, and \citet{feng2025contextual} further illustrate how operational data analytics can enhance performance in contextual newsvendor problems.

Our work is motivated largely by the small-data regime, but the proposed pretrain–finetune framework can also accommodate settings with richer per-instance data in the big-data regime. Pretraining enables the use of expressive models when real observations are limited by providing a strong representation learned from abundant domain-informed data, whereas finetuning adapts that representation to the operational environment using the available real samples. The extent of fine-tuning (i.e., the subset of parameters that are updated) can be chosen to reflect data availability, thereby keeping the approach effective even when proprietary business data are scarce. Consistent with this logic, we show both theoretically and numerically that pretraining can help improve performance, particularly in data-limited settings. On the other hand, when each problem instance is supported by sufficiently many observations (with the number of instances fixed), the setting returns to the big-data regime in which decisions can be learned accurately from the empirical information available for each instance. 
Overall, our framework provides a unified approach that performs well across both regimes.

Conceptually, our work introduces a new LLM-inspired pretrain–then-finetune approach for data-driven stochastic optimization. Rather than deploying off-the-shelf LLMs, we distill the key training principle behind their success and tailor it to our decision-making setting, with accompanying theoretical guarantees. This perspective differs from \citet{huang2025orlm}, which focuses on building open-source LLMs for automated optimization modeling and solution. Several recent studies also draw on Transformer architectures in OM. \citet{wang2023transformer} develop a Transformer model to learn customers' multiple-choice behavior. 
\cite{cao2024probabilistic} emphasize the role of human interaction in AI-enabled systems. \cite{lu2025express} introduce a transformer-based framework to capture customer interactions at both population and individual levels and demonstrate its value for targeting. \cite{hu2025pre} show that pretrained AI models can be used to impute missing covariates and improve decision-making, and \citet{aouad2025sign} study alignment in LLM-based systems in the presence of heterogeneous human preferences. Our problem setting differs substantially from the studies above, and our focus is on how domain-knowledge-guided pretraining and practice-driven fine-tuning interact to jointly contribute to effective learning and optimization. From this perspective, we develop a new methodology grounded in a comprehensive theoretical analysis that clarifies when and why the approach performs well, yielding insights that extend to Transformer-based learning paradigms, including LLMs.

Our work also contributes to another stream of literature on learning theory, focusing on developing theoretical foundations for modern AI models. Despite the striking empirical success of LLMs across domains, rigorous theoretical understanding of these systems, and even of their core architectural primitive, i.e., the Transformer, remains limited. A related body of work studies neural-network learning through the lens of statistical learning theory, characterizing generalization using tools from empirical process theory \citep{anthony1999neural,mohri2018foundations}. More recently, \citet{farrell2021deep} establish nonasymptotic bounds for DNNs and analyze second-stage inference based on the first-step estimation via DNNs. Along this line, \citet{ye2025deep} combine deep learning with doubly robust estimation for causal inference in large-scale experiments and show statistical efficiency, consistency, and asymptotic normality. \citet{guohao} derive minimax-optimal rates for least-squares regression with DNNs and show that intrinsic low-dimensional structure can mitigate the curse of dimensionality. \citet{han2023deep} provide explicit performance guarantees for an end-to-end data-driven newsvendor policy based on deep quantile estimation. \citet{wang2024neural} show that overparameterized neural representations can effectively learn choice models. \citet{cao2024conformal} develops conformal methods with finite-sample guarantees to address model misspecification in data-driven decision-making.

Within this stream, to the best of our knowledge, we provide the first comprehensive theoretical analysis of Transformer-based pretrain-then-finetune learning in relevant settings. The analysis and resulting insights differ significantly from existing theory for standard DNNs. Much of the DNN literature treats the network primarily as a generic function approximator. In contrast, in our framework, the Transformer’s role extends beyond approximation: it is embedded in an LLM-inspired pretrain–then-finetune pipeline tailored to small-data, large-scale optimization. Specifically, pretraining incorporates domain knowledge and supports feasible training even when real data are scarce, while finetuning adapts the model to the operational environment by updating only a targeted subset of parameters, which is both data-efficient and well aligned with the small-data regime. From this standpoint, the pretrain–then-finetune approach itself is central, and the Transformer provides an effective computational vehicle for implementing it. Moreover, our theory explicitly characterizes the interaction between pretraining and fine-tuning, a feature absent from prior analyses and that informs the broader understanding of error behavior in Transformer-based models, such as LLMs. Finally, we quantify how architectural features unique to Transformers, notably the attention mechanism, enter the error decomposition and shape performance through the domain-gap, generalization, and approximation terms. These results, and the accompanying proof techniques, are new to the literature.

\section{Problem Formulation}\label{sec:problem_formulation}

In this paper, we study a small-data, large-scale setting, in which the number of unknown parameter sets is large while the data available to estimate each individual set is limited \citep{mivsic2020data}. Specifically, we consider a firm that must handle a collection of $N$ stochastic optimization problems simultaneously. For instance, the firm may need to make inventory and pricing decisions simultaneously for a large set of products facing random demand \citep{mukherjee2015efficient,miao2022context}.
For each problem $i\in\{1,\ldots, N\}$, the firm chooses a decision $\bx$ and receives a random payoff $c_i(\bx,\xi_i)$, where $\xi_i$ denotes the random outcome. We assume that $\xi_i$ follows a parametric distribution governed by an \emph{unknown} parameter vector $\btheta_i\in\mathbb{R}^p$.
Therefore, the resulting optimization problems are
\begin{equation}\label{eq:formulation}
     \arg\max_{\bx_i\in\mathcal{X}_i}\ v_i(\bx_i,\btheta_i):=\mathbb{E}_{\xi_i\sim\mathbb{P}_{\btheta_i}}[c_i( \bx_i, \xi_i)], \qquad\text{ for }i=1,2,\cdots,N,
\end{equation}
where $\mathbb{P}_{\btheta_i}$ denotes the probability measure governing the random variable $\xi_i$, and its functional form can be flexible;
and $\mathcal{X}_i$ denotes the feasible decision set for task \(i\). While our method and analysis allow for heterogeneity across different problems, we henceforth adopt a common objective $c$, feasible set $\mathcal{X}$, and value function $v$ for notational convenience. 

Although the aggregate amount of data can be substantial when the problem scale $N$ is large, the number of unknown parameters also grows linearly with $N$. As a result, traditional approaches that estimate and optimize each problem in isolation can perform poorly in the small-data regime. At the same time, the task-specific parameters may exhibit shared structure at the population level, which can be leveraged to pool information across tasks and improve both estimation and decision quality \citep{gupta2022data}. To capture and exploit such cross-task connections, we adopt an empirical Bayes (EB) framework \citep{efron2019bayes} and assume that the parameters $\btheta_1,\cdots,\btheta_N$ are independently and identically distributed (i.i.d.) draws from an \emph{unknown} distribution $G$, i.e., $\btheta_i \overset{i.i.d.}{\sim} G$ for  $i=1,\cdots,N$. This assumption is mild and flexible. Practically, it reflects the notion that demand patterns across a large product catalog can be viewed as realizations of an underlying market-level distribution that summarizes common structure, even though that distribution is unobserved. Technically, the EB formulation imposes minimal structure while providing a principled basis for effective transfer learning across tasks.

To implement the optimization in \eqref{eq:formulation}, the firm needs to infer the task-specific parameters from the available observations. To emphasize the small-data nature of the problem, we assume that the firm can access only a single data point for each task when estimating its unknown parameter. This assumption is made without loss of generality and can readily be extended to the case of finitely many observations per task. Specifically, for each $i=1,\ldots,N$, the firm observes $\bd_i\in\mathbb{R}^p$, which we model as a noisy measurement of the unknown parameter $\btheta_i$, capturing measurement error and other sources of deviation. In other words, the observations follow the 
 hierarchical model 
 \begin{equation}\label{generating_process}
     \bd_{i}|\btheta_i \overset{\text{i.i.d.}}{\sim} \mathcal{N}(\btheta_i,\sigma^2 \bI_p),\quad \btheta_i\sim G,\quad \text{ for }i=1,\cdots,N,
 \end{equation}
 where $\mathcal{N}$ denotes the normal distribution, $\sigma$ (assumed known) captures the observation noise level (and thus precision), and $\bI_p$ is the $p\times p$ identity matrix. As before, we impose a common $\sigma$ to streamline notation; accommodating task-specific observation quality is straightforward.
 We want to highlight that this nonparametric formulation is flexible as it allows the shared prior $G$ to be arbitrary rather than restricting it to a parametric family, so the induced marginal distribution $f_G$ for the observations remains sufficiently rich to capture a broad class of data-generating processes.

Effective data-driven decision-making in this setting requires pooling information across tasks and extracting shared structure from the set of observations. Under the EB formulation, this transfer learning process is naturally interpreted as learning the unknown prior distribution $G$. To remain broadly applicable, it is desirable to estimate $G$ without imposing restrictive parametric assumptions, thereby allowing the procedure to adapt flexibly and automatically across different application contexts. A classical approach is nonparametric maximum likelihood estimation (NPMLE); see, e.g., \cite{jiangANDzhang_normalNPMLE,EB-GAUSSIAN-2}. However, NPMLE can be computationally burdensome in high dimensions. More importantly, in small-data regimes where each task contributes only a few and potentially noisy observations, purely data-driven transfer may be statistically fragile, in part because the learning process begins with no informative structure at initialization. Motivated by the success of Transformer-based LLMs, such as the GPT series, which are pretrained on web-scale corpora, we propose to incorporate domain knowledge into the learning and optimization pipeline via a pretraining stage. Crucially, pretraining can be conducted using synthetic data, thereby avoiding reliance on scarce proprietary observations while encoding structural information derived from managerial judgment, stylized models, or prior experience. Starting from this domain-informed initialization, we then finetune the model on real data to adapt it to the operational environment, balancing prior structure with empirical alignment. In the limiting case where domain knowledge perfectly matches practice, finetuning makes little or no adjustment. Conversely, when domain knowledge is imperfect, performance can still improve as additional real data accumulate and finetuning becomes more informative. We describe the proposed LLM-inspired methodology in Section \ref{sec:method}, develop theoretical guarantees in Section \ref{sec:theory}, and validate the approach numerically in Section \ref{sec:numerical}.



\section{A New Pretrain-Then-Finetune Paradigm Inspired by LLM}\label{sec:method}

To solve the target problem, we adopt a two-step estimate-then-optimize (ETO) procedure. In large-scale settings, ETO offers greater flexibility than integrated estimate-and-optimize methods for applying estimation to various downstream tasks. Once the unknown parameters are estimated, each stochastic optimization problem reduces to a deterministic one and can then be solved independently. Importantly, although the optimization step is task-specific, the estimation stage incorporates cross-task information sharing; as a result, transfer learning effects induced during estimation naturally carry over to downstream decisions. Accordingly, we focus this section on the estimation component of the procedure. We adopt an LLM-inspired pretrain-then-finetune paradigm to obtain efficient and scalable parameter estimates; see Section \ref{subsec:our_method} for details.


Our method builds on a designed Transformer model, the core architectural primitive underlying modern LLMs. Compared with standard architectures such as DNNs and recurrent neural networks (RNNs), Transformers offer advantages that extend beyond generic function approximation and are particularly well aligned with the pretrain–finetune paradigm. On one hand, they have demonstrated strong generalization and transferability in practice: after pretraining on massive raw data (e.g., web-scale text corpora), an LLM can often perform well on previously unseen tasks through prompting alone, even without updating model parameters \citep{ICL1}. On the other hand, Transformers can be adapted effectively to new tasks using comparatively small amounts of task-specific data and computation \citep{PEFT1}, yielding substantial improvements in downstream performance. This capability is further enhanced by parameter-efficient finetuning methods such as low-rank adaptation (LoRA; \citealt{hu2022lora}), which update only a selected subset of parameters while keeping the remainder fixed, thereby reducing both data requirements and computational cost. These properties are particularly attractive in the small-data, large-scale optimization setting we study. We next introduce the Transformer architecture used in our framework in Section \ref{subsec:transformer_architecture}.

\subsection{Preliminary: Transformer Architecture}\label{subsec:transformer_architecture}

\begin{figure}
    \centering
    \includegraphics[width=0.96\linewidth]{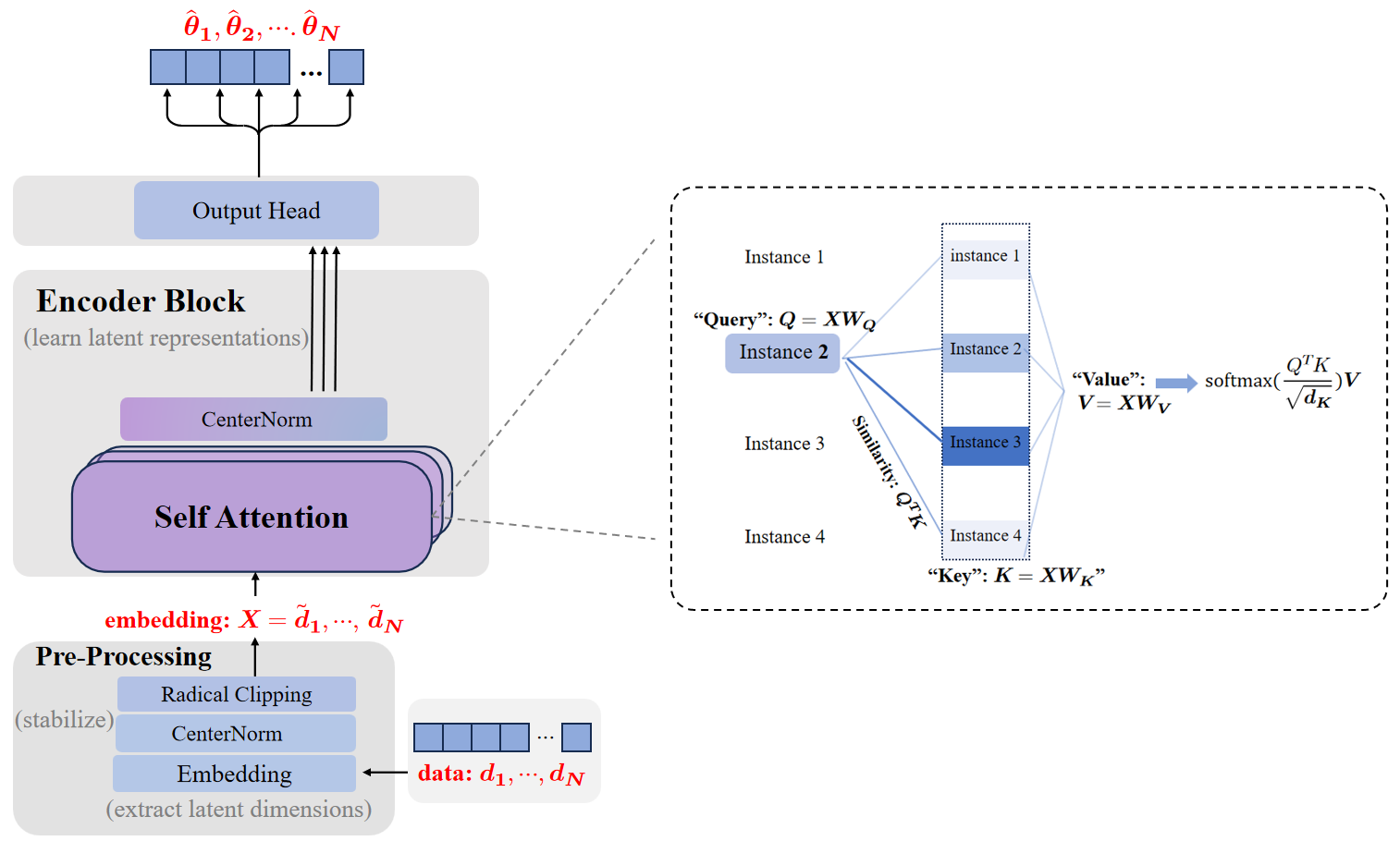}
    \caption{The architecture of our adopted Transformer model.}
    \label{fig:EBTF}
\end{figure}

At a high level, the Transformer is a foundational neural architecture that maps input sequences to output representations by learning attention-based dependencies across elements. In our context, we design a specific encoder-only Transformer, illustrated in Figure~\ref{fig:EBTF}.\footnote{Many Transformer-based variants, such as the GPT family, employ decoder-only or encoder–decoder architectures, where the decoder is primarily designed for autoregressive content generation. By contrast, the encoder is designed to process input data and produce representations that capture latent structure. Because our objective is parameter estimation rather than sequence generation, an encoder-only architecture is particularly well suited to our application.} The model takes as input a sequence of data whose length may vary and outputs corresponding parameter estimates. Formally, we can view the Transformer as a mapping from observations $\bbD=(\bd_1,\cdots,\bd_N)\in (\mathbb{R}^{p})^{N}$ to estimates $\hat{\btheta}=(\hat{\btheta}_1, \dots, \hat{\btheta}_N)\in (\mathbb{R}^{p})^{N}$, denoted by $\mathcal{T}_{\textsf{TF}}(\cdot;W): (\mathbb{R}^{p})^{N}\to (\mathbb{R}^{p})^{N}$, where $W\in\Phi$ represents the model parameters and $\Phi$ denotes the parameter space. As depicted in Figure~\ref{fig:EBTF}, our Transformer architecture consists of three main modules, arranged in data-flow order: a pre-processing block, an encoder block, and an output head. Drawing an analogy to natural language processing (NLP), each observation $\bd_i$ can be viewed as a \textit{token} in the input sequence, analogous to a word token in text. We describe each component of the architecture in detail below.

\textbf{Pre-Processing Block.} This block prepares the raw input sequence for the encoder, and it consists of three major steps; see Figure \ref{fig:EBTF}. The first is an embedding layer. In NLP, embedding layers map discrete tokens to continuous vectors, enabling their processing within a unified framework. Analogously, in our setting, each observation $\bd_i$ is passed through an embedding network, a multilayer perceptron (MLP) with ReLU activations, to produce an embedding $\Tilde\bd_i \in \mathbb{R}^{p_{\mathrm{emb}}}$.
We denote the depth (number of hidden layers) and width (maximum number of neurons per hidden layer) of this MLP by $\mathcal{D}_{\mathrm{emb}}$ and $\mathcal{W}_{\mathrm{emb}}$, respectively. Conceptually, this step extracts lower-dimensional latent representations that facilitate subsequent representation learning in the encoder. 
The resulting embeddings are then processed sequentially by a normalization layer (referred to as CenterNorm in Figure \ref{fig:EBTF}) and a radical clipping layer that enforces uniform bounds. These operations improve numerical stability during training and serve an additional role in the theoretical analysis by regularizing the embedding space; in particular, they help ensure that the Transformer’s output is Lipschitz continuous with respect to its input (see the online appendix for details).


\textbf{Encoder Block.} The encoder block implements the self-attention mechanism, which is the defining architectural feature of Transformers. Its role is to learn representations from input embeddings that capture shared structure and cross-token interactions, including interactions across tasks. In NLP, the term ``attention” refers to relevance weights assigned to other words in a sentence when handling a given word. Analogously, in our setting, attention weights quantify how information from other tasks contributes to estimating the parameter of the currently focused task. This mechanism thus provides an explicit and quantitative channel for transfer learning: embeddings deemed more relevant receive higher attention weights and exert greater influence on the resulting representation and parameter estimate.

\begin{definition}[Self-Attention Mechanism]\label{def:attention1}
For matrices $\bW_Q, \bW_K \in \mathbb{R}^{ p_\mathrm{emb}\times p_K } $ and $\bW_V \in \mathbb{R}^{  p_\mathrm{emb}\times p_V}$ and an embedding sequence $\Tilde{\bbD}:=(\Tilde\bd_1,\ldots,\Tilde\bd_N)^\top\in\mathbb{R}^{N\times p_{\mathrm{emb}}}$, the \emph{single-head self-attention} mapping $f$ is defined as
\begin{equation}\label{eq:attention:1}
f(\Tilde{\bbD})=\mathrm{softmax}\left(\frac{\bQ \bK^\top}{\sqrt{p_K}}\right)\bV\in \mathbb{R}^{N\times p_V},
\end{equation}
where $\bQ:=\Tilde{\bbD} \bW_Q, \bK:=\Tilde{\bbD} \bW_K \in \mathbb{R}^{N \times p_K}$, and $\bV:=\Tilde{\bbD} \bW_V\in \mathbb{R}^{N \times p_V}$; the softmax function $\mathrm{softmax}(\cdot)$ is applied to each row of $\bQ\bK^\top/\sqrt{p_K}$, i.e., the $(i,j)$-th entry of $\mathrm{softmax}\left({\bQ\bK^\top}/{\sqrt{p_K}}\right)$ is
\begin{equation}\label{eq:attention_2}
    \omega_{ij}
:=
\frac{\exp\left(\bQ_i \bK_j^\top / \sqrt{p_K}\right)}
{\sum_{j'=1}^{N} \exp\left(\bQ_i \bK_{j'}^\top / \sqrt{p_K}\right)},
\end{equation}
 where $\bQ_i$, $\bK_j$, and $\bV_j$ denote the $i$-th, $j$-th, and $j$-th rows of 
$\bQ$, $\bK$, and $\bV$, respectively. We further define the $i$-th row of $f(\Tilde{\bbD})$ as $f_i(\Tilde{\bbD}) = \sum_{j=1}^{N}\omega_{ij} \bV_j$, and refer to the matrices $\bW_Q, \bW_K,$ and $\bW_V$ as the parameters of the attention mechanism.
\end{definition}

From Definition~\ref{def:attention1} and the right panel of Figure~\ref{fig:EBTF}, 
each embedding element $\Tilde\bd_i$ enters the attention mechanism in three distinct roles: 
\begin{itemize}
    \item Query. When task $i$ is the focal instance, $\Tilde\bd_i$ is mapped to a query vector used to compare against all other embeddings. Specifically, in \eqref{eq:attention:1}, the $i$-th row of the query matrix $\bQ := \Tilde{\bbD} \bW_Q $ represents the query associated with task $i$, and the parameter matrix $\bW_Q$ governs how each embedding is transformed into a query representation.
    
    \item Key. Each embedding $\Tilde\bd_i$ also serves as a comparison object when processing other embeddings $\Tilde\bd_j$ for $j\neq i$. In \eqref{eq:attention:1}, the $j$-th row of the key matrix $\bK := \Tilde{\bbD} \bW_K$ represents the key associated with task $j$, and the parameter matrix $\bW_K$ aggregates the corresponding effect via keys. The attention mechanism then evaluates the similarity between the focal task $i$ and every task $j\in\{1,\ldots,N\}$ via the dot product, i.e., $\bQ_i \bK_j^\top$, 
    and applies a row-wise softmax normalization to obtain the attention weights $\bomega_i = 
(\omega_{i1}, \omega_{i2}, \ldots, \omega_{iN})^\top$ in \eqref{eq:attention_2}.
    
    \item Value. Values carry information aggregated according to attention weights. Specifically, the $j$-th row of the value matrix $\bV := \Tilde{\bbD} \bW_V $ represents the value representation for task $j$, parameterized by $\bW_V$. The resulting representation for task $i$ in the encoder block is the weighted sum $f_i(\Tilde{\bbD}) = \sum_{j=1}^{N}\omega_{ij} \bV_j$, where $\omega_{ij}$, computed from the query-key similarities, quantifies the contribution of the task $j$'s value $\bV_j$ to the representation of task \(i\).
\end{itemize}
Compared with alternative AI models such as DNNs and RNNs, attention mechanisms have been shown to deliver strong performance, particularly in capturing long-range dependencies in LLM applications. This capability is especially valuable in large-scale settings, where each instance may benefit from leveraging information dispersed across a large collection of related tasks.
 
Beyond single-head attention, an extended {multi-head attention} mechanism can be employed, in which multiple attention heads operate in parallel. Each head can be viewed as a separate self-attention module that attends to the input sequence through a distinct projection, and the collection of heads enables the model to capture heterogeneous interaction patterns among input instances \citep{lu2025express}. The head-specific representations are then concatenated to form the final encoder representation, 
thereby enhancing the ability to model diverse dependencies. We defer the formal definition of multi-head attention to the online appendix.
Finally, at the end of the encoder block in Figure~\ref{fig:EBTF}, we apply an additional normalization layer to further stabilize training. 
    
\textbf{Output Head.} The last component in Figure \ref{fig:EBTF} is an output head implemented as an MLP with ReLU activations,  of depth $\mathcal{D}_{\mathrm{oh}}$ and width $\mathcal{W}_{\mathrm{oh}}$. It produces the final point estimates $\hat{\btheta}=(\hat{\btheta}_1, \dots, \hat{\btheta}_N)\in\mathbb{R}^{p\times N}$. 
The output head serves two roles. 
First, the ReLU nonlinearities increase the model’s expressive capacity. Second, the MLP maps the encoder’s latent representations into the parameter space, ensuring dimensional consistency and yielding interpretable parameter estimates.


\subsection{Our Pretrain-Then-Finetune Method}\label{subsec:our_method}

\begin{figure}[h!]
    \centering
    \includegraphics[width=0.96\linewidth]{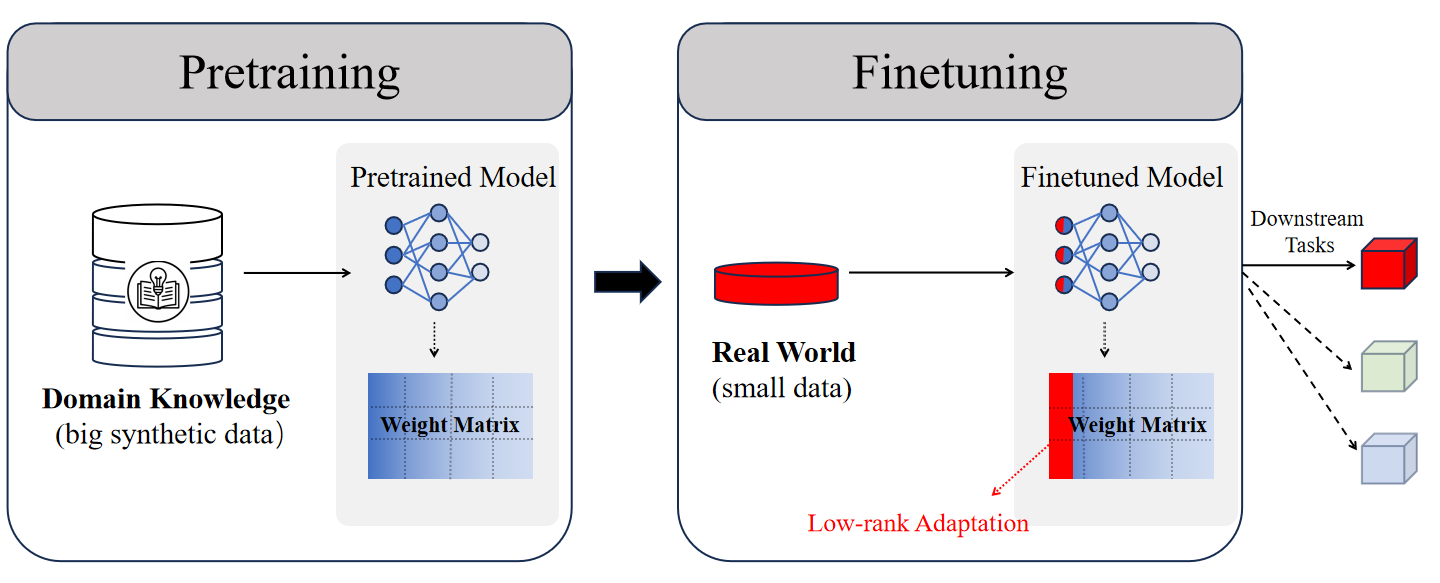}
    \caption{Illustration of our pretrain-then-finetune framework.}
    \label{fig:prefine}
\end{figure}

The central difficulty in the considered small-data, large-scale problem \eqref{eq:formulation} is estimating a large set of task-specific parameters simultaneously from only sparse observations per task. This challenge is particularly acute for machine learning tools, which typically require substantial data to train reliably from scratch. To address this issue, we adopt an LLM-inspired \textit{pretrain-then-finetune} strategy, illustrated in Figure~\ref{fig:prefine}. We first pretrain the model on a large corpus of relevant historical data (if available) and on domain-informed synthetic data designed to encode structural knowledge about the application. During pretraining, the Transformer learns generic latent representations that embed domain expertise and provide an informative initialization for subsequent estimation. In the fine-tuning stage, we adapt the pretrained model to the target environment using limited real-world observations. To remain effective in the small-data regime, we fix most parameters and update only a subset, thereby reducing both data requirements and computational cost while improving alignment with the task distribution. Importantly, neither pretraining nor finetuning is a direct application of standard routines. We introduce several problem-specific design choices to tailor the procedure to our setting, as detailed below.

\subsubsection{Supervised Pretraining.}\label{subsubsec:Supervised Pretraining}
As described above, the Transformer is pretrained on a collection of synthetic datasets of size $K$, denoted by $\{{\bbD}^{\mathrm{pre}}_k\}_{k=1}^K$. Each pretraining instance $\bbD^{\mathrm{pre}}_k:=(\bd_{k,1}, \dots, \bd_{k,N_k})$ is a sequence that may vary in length \(N_k\). Without loss of generality, we assume that the pretraining data are generated according to
\begin{align*}
    \bd_{k,i} \sim f(\cdot \mid \btheta^{\mathrm{pre}}_{k,i}),\quad \btheta^{\mathrm{pre}}_{k,i} \sim G^{\mathrm{pre}},\quad\text{for } k=1,\ldots,K \text{ and } i=1,\cdots,N_k.
\end{align*}
In other words, the pretraining data follow the same hierarchical structure as in \eqref{generating_process}. There are, however, two important distinctions. First, the latent parameters ${\btheta}_k^{\mathrm{pre}}:=(\btheta^{\mathrm{pre}}_{k,1},\cdots,\btheta^{\mathrm{pre}}_{k,N_k})$ are known for the pretraining datasets, which enables supervised learning. Second, the prior distribution $G^{\mathrm{pre}}$ is specified by the firm and reflects domain knowledge about the problem context. This prior may incorporate managerial experience, insights from theoretical models in the literature, or empirical regularities drawn from related historical settings. As we show theoretically in Section~\ref{sec:theo:1}, informative domain knowledge can substantially improve performance.


During pretraining, each sequence ${\bbD}^{\mathrm{pre}}_k$ is fed into the Transformer, which outputs the corresponding parameter estimates
$$\mathcal{T}_{\textsf{TF}}(\bbD^\mathrm{pre}_k;W)=\hat{\btheta}_k^\mathrm{pre}=(\hat\btheta^{\mathrm{pre}}_{k,1},\cdots,\hat\btheta^{\mathrm{pre}}_{k,N_k}),$$
where $W$ denotes the model parameters.
The pretraining objective minimizes the mean squared error (MSE) between the model's outputs and the corresponding true labels ${\btheta}_k^{\mathrm{pre}}$, i.e.,
\begin{equation}\label{pretaining_loss}
W_0:=\argmin_{W\in\Phi}~ \mathcal{L}^{\mathrm{pre}}(\mathcal{T}_{\textsf{TF}}) := \sum_{k=1}^K \left\|\mathcal{T}_{\textsf{TF}}(\bbD^\mathrm{pre}_k;W)- {\btheta}_k^{\mathrm{pre}} \right\|^2,
\end{equation}
where $\|\cdot\|$ denotes the $\ell_2$ norm and $\Phi$ is the parameter space.
We denote the resulting pretrained model by $\mathcal{T}_{\textsf{TF}}(\cdot;W_0)$.

\subsubsection{Label-Free Finetuning.}\label{subsubsec:finetuning}
Although pretraining can leverage abundant data, it serves only as an initialization for the estimation tasks of interest. We therefore finetune the pretrained model using the observed task-specific dataset ${\bbD} = (\bd_1, \dots, \bd_N)$. Unlike pretraining, however, fine-tuning faces a fundamental challenge: ground-truth labels (i.e., the true parameter vectors) are unavailable in real-world applications, so a direct supervised loss is also unavailable. Concretely, when $\bbD$ is fed into $\mathcal{T}_{\textsf{TF}}$ and the model outputs the estimates $\mathcal{T}_{\textsf{TF}}({\bbD}) = (\hat{\btheta}_1, \dots, \hat{\btheta}_N)$, the corresponding true parameters $\btheta_1, \dots, \btheta_N$ are unavailable. Consequently, standard comparison-based objectives, such as MSE, cannot be computed to guide fine-tuning.

To address the absence of labels and obtain a tractable finetuning objective, we invoke a generalized Stein identity to derive a loss function that eliminates dependence on the unobserved parameters. Specifically, the MSE of any candidate finetuning model $\mathcal{T}_{\textsf{TF}}$ can be rewritten as
\begin{equation}\label{eq:ERM_fmodeling}
  \mathcal{L}(\mathcal{T}_{\textsf{TF}}):=  \mathbb{E}_{\btheta\sim G,\bd\sim f(\cdot|\theta)}\left[   \|\mathcal{T}_{\textsf{TF}}(\bd)-\btheta\|^2\right]\overset{\substack{\text{Stein's}\\\text{identity}}}{=}\quad\underbrace{\mathbb{E}_{\bd\sim f_G} \left[ \|\mathcal{T}_{\textsf{TF}}(\bd)\|^2 - 2 d^\top \mathcal{T}_{\textsf{TF}}(\bd) + 2 \sigma^2 \nabla \cdot \mathcal{T}_{\textsf{TF}}(\bd) \right]}_{\mathcal{L}^{\mathrm{finetune}}(\mathcal{T}_{\textsf{TF}})},
\end{equation}
where $\nabla \cdot \mathcal{T}_{\textsf{TF}}$ denotes the divergence of the mapping $\mathcal{T}_{\textsf{TF}}$. The proof of the equality is provided in the online appendix. We denote the right-hand side by $\mathcal{L}^{\mathrm{finetune}}(\mathcal{T}_{\textsf{TF}})$. This objective depends only on the observed data and the model outputs, is directly computable, and is exactly equivalent to the MSE. To this end, the finetuning stage proceeds by empirical risk minimization (ERM) based on the expected loss in \eqref{eq:ERM_fmodeling}. Specifically, we solve
\begin{equation}\label{eq:finetune_loss_emp}
    \argmin_{\Delta W:W_0+\Delta W\in\Phi}\hat{\mathcal{L}}^{\mathrm{finetune}}_N(\mathcal{T}_{\textsf{TF}}) := \frac{1}{N} \sum_{i=1}^N \left[ \|\mathcal{T}_{\textsf{TF}}(\bd_i)\|^2 - 2 \bd_i^\top \mathcal{T}_{\textsf{TF}}(\bd_i) + 2 \sigma^2 \nabla \cdot \mathcal{T}_{\textsf{TF}}(\bd_i) \right].
\end{equation}
It is worth emphasizing that, given the pretrained parameters $W_0$, finetuning retrains only a small subset of the Transformer parameters through $\Delta W$. Specifically, we adopt low-rank adaptation (LoRA), a widely used parameter-efficient finetuning approach in LLMs. As we briefly describe below, LoRA incorporates two key ideas that make finetuning feasible in the small-data regime.

{First}, adapting a pretrained Transformer does not require updating all components. Instead, one can finetune only selected modules, such as the embedding layer, the output head, or a subset of weights within the attention block, while keeping the majority of the model fixed. In our experiments, for example, updating only the embedding layers and the output head is sufficient, with the attention block left frozen. This finding suggests that attention patterns learned during pretraining on synthetic data already capture transferable structure that remains valuable when adapting to real-world tasks. {Second}, even for modules that are fine-tuned, LoRA restricts updates to a \emph{low-rank} modification of the pretrained weights. To illustrate, with a slight abuse of notation, let $W_0 \in \mathbb{R}^{d_{\mathrm{out}}\times d_{\mathrm{in}}}$ denote the pretrained weight matrix of a module selected for finetuning (e.g., a weight matrix in the output head), where $d_{\mathrm{in}}$ and $d_{\mathrm{out}}$ are its input and output dimensions.
LoRA writes the adapted weight matrix as 
\begin{equation}\label{eq:lora_1}
    W =\underbrace{ W_{0}}_{\mathrm{pretrained\ model}} + \underbrace{\Delta W}_{\mathrm{finetuning\ adaptation}},
\end{equation}
with $$\Delta W=BA,$$
where $B \in \mathbb{R}^{d_{\mathrm{out}}\times r}$ and $A \in \mathbb{R}^{r\times d_{\mathrm{in}}}$. 
 Importantly, here $r \ll \min\{d_{\mathrm{out}}, d_{\mathrm{in}}\}$ is a prescribed rank, typically chosen to be small in practice.  
Finetuning updates only $A$ and $B$, while the pretrained backbone $W_0$ remains fixed. This parameterization constrains the update to a low-dimensional subspace: the number of trainable parameters is $2r(d_{\mathrm{out}} + d_{\mathrm{in}})$, rather than $d_{\mathrm{out}}d_{\mathrm{in}}$. Consequently, the model learns an effective adaptation direction without re-optimizing the entire weight matrix, substantially reducing both computational burden and the amount of real data required.


\section{Theoretical Guarantees for Our Transformer-Based Learning and Optimization}\label{sec:theory}

In this section, we establish theoretical guarantees for our pretrain-then-finetune approach introduced in Section \ref{sec:method}. To the best of our knowledge, our results provide the first comprehensive theoretical analysis of Transformer learning in relevant settings and offer a principled account of why the pretrain–finetune paradigm can be effective in LLM-style pipelines. More broadly, the analysis clarifies when and why Transformer-based models perform well, with implications for LLMs built on the same architectural backbone. Specifically, Section~\ref{sec:theo:1} presents a general decomposition of the estimation error, yielding high-level insights into when pretraining is critical and when finetuning alone can be sufficient. Section~\ref{sec:theo:2} then bounds each error component and characterizes how they depend on key problem and model primitives, including the number of tasks, architectural features of the Transformer, and the quality of domain knowledge encoded in pretraining. Finally, Section~\ref{sec:final} connects estimation error to decision performance under the Transformer-based ETO framework for \eqref{eq:formulation}.

\subsection{High-Level Insights: When Pretraining is Critical and When Finetuning Suffices}
\label{sec:theo:1}
Estimation accuracy is central to downstream decision quality: under the ETO framework, estimation error propagates directly into the eventual decision error. To evaluate the LLM-inspired estimation method introduced in the previous section, we first present a high-level error decomposition that yields clear insights into the respective roles of pretraining and finetuning. We first introduce several common performance metrics. For a generic parameter estimator $\mathcal{T}(\cdot): \mathbb{R}^p \to \mathbb{R}^p$ that maps an observation $\bd\in\mathbb{R}^p$ to an estimate of the corresponding task parameter, we measure estimation performance by the MSE:
\begin{equation}\label{eq:square_loss}
   \mathcal{L}(\mathcal{T}) := \mathbb{E}_{\bd|\btheta\sim f(\cdot|\btheta),\ \btheta\sim G} \left[ \| \mathcal{T}(\bd) - \btheta \|^2 \right].
\end{equation}
That is, the MSE is evaluated on a new,  independent draw from the hierarchical model in \eqref{generating_process}. In what follows, when the context is clear, we will omit subscripts in the expectation operator $\mathbb{E}$ for brevity.
If the prior $G$ were known, the \emph{oracle} benchmark that minimizes \eqref{eq:square_loss} would be the posterior mean, i.e., $\mathcal{T}^\ast(\bd):=\mathbb{E}[\btheta|\bd]$. 
Accordingly, for any data-driven estimator $\mathcal{T}$ constructed from the available data $\mathscr{D}$, we define its \emph{estimation error} as the excess MSE relative to the oracle:
\begin{equation}\label{eq:excess_risk_est}
    \mathcal{E}_\mathrm{est}(\mathcal{T}) :=\mathbb{E}_{\bbD}[ \mathcal{L}(\mathcal{T}) - \mathcal{L}(\mathcal{T}^*)]=\mathbb{E}_{\bbD}\mathbb{E}_{\bd\sim f_G} \left[ \| \mathcal{T}(\bd) -\mathcal{T}^\ast(\bd) \|^2 \right],
\end{equation}
where the second equality follows from the orthogonality principle for the posterior mean $\mathcal{T}^\ast$. The outer expectation over $\bbD$ accounts for sampling variability: since $\bbD$ is random, the resulting estimator is random as well.

\begin{figure}
    \centering
    \includegraphics[width=0.69\linewidth]{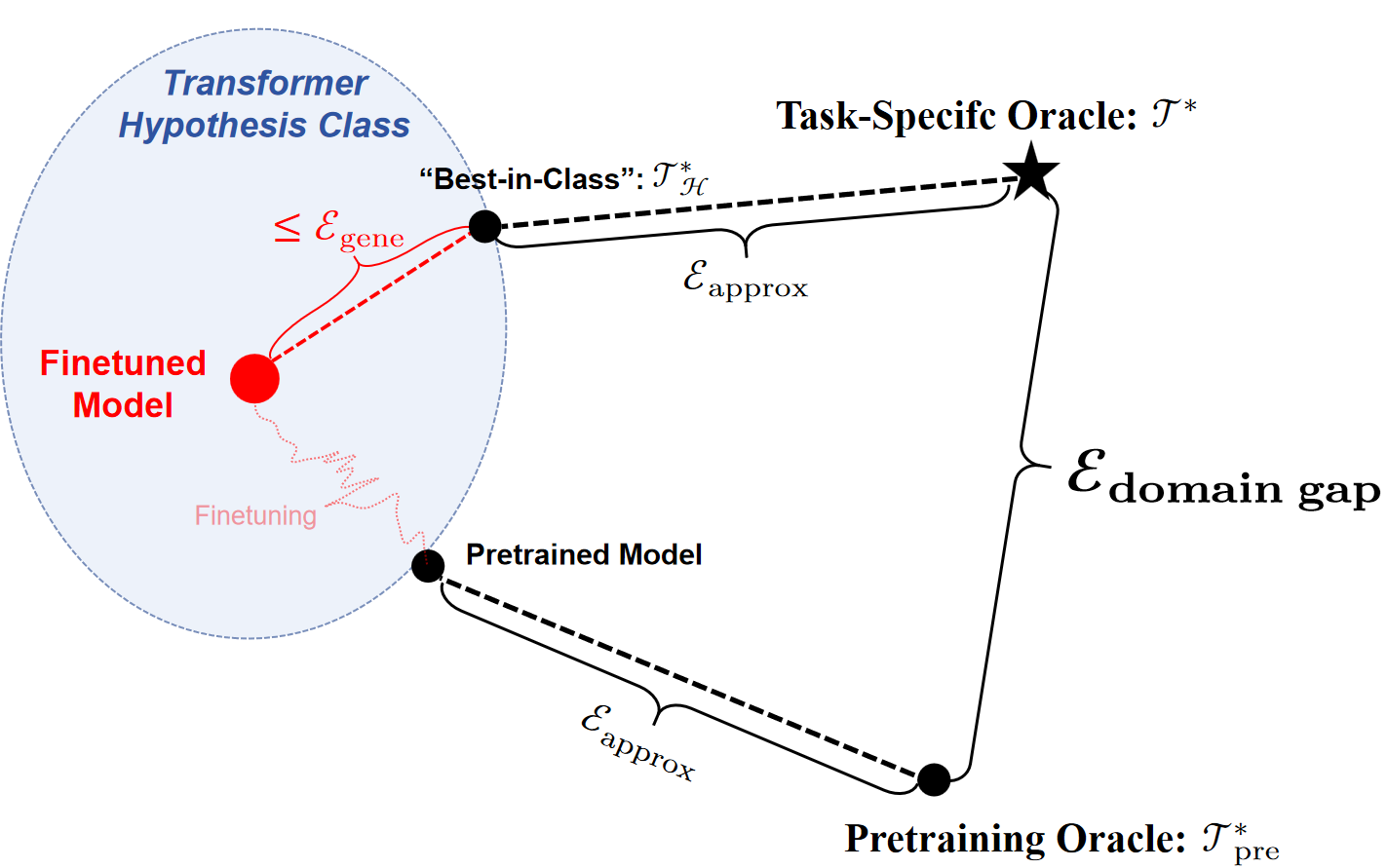}
    \caption{Overview of our pretraining–finetuning error analysis framework.}
    \label{fig:overview}
\end{figure}

Let $\hat{\mathcal{T}}_{\textsf{TF}}$ denote the Transformer estimator obtained by pretraining on the synthetic datasets and subsequently finetuning on the real sample $\bbD$. Intuitively, its estimation error may stem from (i) the finite expressive power of the chosen Transformer architecture, (ii) statistical error induced by finetuning with limited and noisy observations, and (iii) potential mismatch between the pretraining distribution and the true data-generating environment. To formalize these effects and highlight the interaction between pretraining and finetuning, we decompose the estimation error into three components: an \emph{approximation error}, a \emph{finetuning generalization error}, and a \emph{domain gap}; see Figure~\ref{fig:overview} for an illustrative overview.

The first source of error arises from the model’s finite representational capacity. For a given architecture, the Transformer ranges only over a restricted hypothesis class mapping data to estimates (the shaded ellipse in Figure~\ref{fig:overview}), defined as $\mathcal{H}:=\{\mathcal{T}_{\textsf{TF}}(\cdot;W):W\in\Phi\}$. The oracle mapping $\mathcal{T}^\ast$ needs not belong to $\mathcal{H}$, and this misspecification induces an \emph{approximation error} defined as\footnote{For complete rigor, there is also an approximation error in the pretraining stage relative to the 
pretraining target $\mathcal{T}^\ast_{\mathrm{pre}}$ (introduced in \eqref{def:pre:task}), as shown in Figure \ref{fig:overview}. However, $\mathcal{T}^\ast$ and $\mathcal{T}^\ast_{\mathrm{pre}}$ differ only through the prior distribution used in the corresponding Bayes estimators and thus share the same functional form.
As a result, the approximation analysis developed below applies to either target with only notational changes. For clarity, we only state definitions and results for $\mathcal{E}_{\mathrm{approx}}$ with respect to 
$\mathcal{T}^\ast$.}
    \begin{equation}
        \label{def:approx}\mathcal{E}_{\mathrm{approx}}:= \mathcal{L}(\mathcal{T}_{\mathcal{H}}^\ast)-\mathcal{L}(\mathcal{T}^\ast),
    \end{equation}
where $\mathcal{T}_{\mathcal{H}}^\ast 
:= 
\arg\min_{\mathcal{T} \in \mathcal{H}} \mathcal{L}(\mathcal{T})$ is the \emph{best-in-class} estimator within $\mathcal{H}$. 
As illustrated in Figure~\ref{fig:overview}, $\mathcal{T}_{\mathcal{H}}^\ast$ can be interpreted as the projection of the target mapping $\mathcal{T}^\ast$ onto the hypothesis space $\mathcal{H}$, and $\mathcal{E}_{\mathrm{approx}}$ quantifies the resulting discrepancy. As we discuss in Section~\ref{sec:theo:2}, the approximation error is governed by the Transformer’s representational capacity, including the depth and width of the MLP components and the dimensions of the attention mechanism (e.g., $p_K$ and $p_V$). All else equal, larger models typically yield greater expressiveness and therefore smaller approximation error.

The second source of error is the \emph{finetuning generalization error}. It is incurred because the finetuned estimator $\hat{\mathcal{T}}_{\textsf{TF}}$ is obtained by minimizing the empirical objective $\hat{\mathcal{L}}^{\mathrm{finetune}}_N(\mathcal{T}_{\textsf{TF}})$ on the finite sample $\bbD$ (see \eqref{eq:finetune_loss_emp}), rather than the population loss $\mathcal{L}^{\mathrm{finetune}}(\mathcal{T}_{\textsf{TF}})$ in~\eqref{eq:ERM_fmodeling}. Sampling noise in $\bbD$ can therefore lead $\hat{\mathcal{T}}_{\textsf{TF}}$ to deviate from the population-optimal solution within the model class. We quantify this effect by the gap between the expected and empirical finetuning objectives evaluated at the learned estimator:
$$\mathcal{E}_{\mathrm{gene}}
:=\mathbb{E}_{\bbD}[\mathcal{L}^{\mathrm{finetune}}(\hat{\mathcal{T}}_{\textsf{TF}}) - \hat{\mathcal{L}}^{\mathrm{finetune}}_N(\hat{\mathcal{T}}_{\textsf{TF}})].$$
Here, the expectation is taken over the randomness in $\bbD$ since both $\hat{\mathcal{L}}^{\mathrm{finetune}}_N$ and $\hat{\mathcal{T}}_{\textsf{TF}}$ depend on the realized sample. As $N$ increases, the empirical objective concentrates around its population counterpart, and $\mathcal{E}_{\mathrm{gene}}$ is expected to decrease. We provide a detailed analysis in Section~\ref{sec:theo:2}.

The third component, central and new to our LLM-inspired approach, captures the quality of pretraining. Recall that in Section \ref{subsec:our_method}, we pretrain the Transformer on synthetic datasets to obtain an informative initialization for the real tasks. Because domain knowledge is imperfect, the pretraining marginal distribution $f_{G^{\mathrm{pre}}}$ used in this stage may differ from the true data-generating distribution $f_G$. Consequently, pretraining can steer the model toward a different target mapping (illustrated in Figure~\ref{fig:overview}), namely the Bayes oracle associated with the pretraining distribution rather than the oracle for the true environment. This phenomenon mirrors a well-known limitation of general-purpose LLMs: when the pretraining distribution is poorly aligned with a specialized task domain, e.g., short-horizon stock-price prediction, performance can be unreliable because the pretrained representations may be systematically miscalibrated for the target setting.
Mathematically, the target 
mapping induced by the pretraining stage is 
\begin{equation}\label{def:pre:task}
    \mathcal{T}_{\mathrm{pre}}^\ast
:= \arg\min_{\mathcal{T}} 
\mathbb{E}_{\bd'|\btheta'\sim f(\cdot|\btheta'),\ \btheta'\sim G^{\mathrm{pre}}}
\big[ \| \mathcal{T}(\bd') - \btheta' \|^2 \big].
\end{equation}
This population objective corresponds to the supervised pretraining loss in \eqref{pretaining_loss} in Section \ref{subsubsec:Supervised Pretraining}, where we know the latent parameters $\btheta'$. In general, $\mathcal{T}_{\mathrm{pre}}^\ast$ may not coincide with the target oracle $\mathcal{T}^{\ast}$, unless $G^{\mathrm{pre}}=G$. We refer to the resulting discrepancy as the \emph{domain gap}, defined by
$$
\mathcal{E}_{\mathrm{domain}}
:= \mathcal{L}(\mathcal{T}^{\ast}_{\mathrm{pre}}) - \mathcal{L}(\mathcal{T}^{\ast})
= \mathbb{E}_{\bd\sim f_G}\big[\|\mathcal{T}^\ast_{\mathrm{pre}}(\bd) - \mathcal{T}^\ast(\bd)\|^2\big],
$$
where the final equality again follows from the orthogonality principle for $\mathcal{T}^\ast$.

Taken together, these three components jointly determine estimation accuracy and interact in nontrivial ways, as formalized in the following Proposition~\ref{theo:interaction}. Before that, we introduce two mild technical conditions that ensure that the estimation problem is well-defined.
\begin{assumption}\label{ass:lora}
Let $\mathcal{L}(\mathcal{T}_{\textsf{TF}}(\cdot; W))$ denote the MSE of $\mathcal{T}_{\textsf{TF}}$ with parameter set $W$.
We assume that $\mathcal{L}$ is differentiable with respect to $W$ at the
pretrained parameter $W_0$ defined in \eqref{pretaining_loss}.
\end{assumption}


\begin{assumption}\label{ass:tail}
  Let $g(x)$ and $g^\mathrm{pre}(x)$ be the probability densities of $G$ and $G^\mathrm{pre}$, respectively. There exist constants $c_1,c_2,c_3>0$ such that, for all $\bd\in\mathbb{R}^p$ satisfying $\|\bd\|>c_1$,
 the densities satisfy 
$g(\bd),g^\mathrm{pre}(\bd)\le c_2 e^{-c_3\|\bd\|^2}.$
\end{assumption}


Assumption~\ref{ass:lora} is mild.
Training Transformer and other AI models is typically carried out via gradient-based optimization, for which differentiability of the loss with respect to the model parameters is a standard regularity condition.
Assumption~\ref{ass:tail} imposes mild tail regularity on both $G$ and $G^{\mathrm{pre}}$. It covers common cases in theory and practice, including priors with compact support and, more generally, distributions with tails no heavier than sub-Gaussian \citep{wangmengdi}.

\begin{proposition}[Pretraining-Finetuning Interaction]\label{theo:interaction}
Suppose Assumption~\ref{ass:lora} holds.  
The estimation error satisfies
\begin{equation}\label{eq:excess_risk_decomp}
\mathcal{E}_{\mathrm{est}}
\le
\min\big\{
\underbrace{\mathcal{E}_{\mathrm{gene}}}_{\mathrm{Finetuning}},
 \quad
\underbrace{\mathcal{E}_{\mathrm{domain}}}_{\mathrm{Pretraining}}
\big\}
+\mathcal{E}_{\mathrm{approx}}.
\end{equation}

\end{proposition}

Proposition~\ref{theo:interaction} provides the high-level insights into how pretraining and finetuning jointly shape estimation accuracy. The approximation error constrained by the chosen Transformer architecture may persist regardless of the training pipeline. Conditional on this architectural limit, however, the pretrain-then-finetune approach can effectively leverage whichever signal is stronger: a small finetuning generalization error or a small domain gap. In other words, the bound in \eqref{eq:excess_risk_decomp} is driven by the better of the two. This mechanism highlights an important advantage over purely empirical approaches that rely exclusively on task-specific real data. Such methods are governed solely by the finetuning component in \eqref{eq:excess_risk_decomp} and cannot exploit additional information available through relevant historical data or structured domain knowledge. By contrast, the pretrain–finetune paradigm can incorporate these sources through pretraining, a practice central to the success of modern LLMs. In small-data decision environments, where real observations are scarce and costly, Proposition~\ref{theo:interaction} implies that informative pretraining can even dominate finetuning and materially improve estimation accuracy. 
Our numerical results in Section~\ref{sec:numerical} further suggest that pretraining can improve performance even when the finetuning protocol is held fixed, indicating benefits beyond the conservative ``either-or” structure captured by \eqref{eq:excess_risk_decomp}.

Proposition~\ref{theo:interaction} also yields a transparent regime interpretation. When real data are scarce---much as when one queries an LLM for a specialized task with limited supporting information---the estimator must rely primarily on the pretrained model and thus on the domain knowledge encoded during pretraining. In this regime, the finetuning generalization error can exceed the domain gap, so the estimation error is driven mainly by the latter. The approximation error can be made small with sufficient model capacity; we formalize this in Theorem~\ref{theo:approx}. As shown in Section~\ref{sec:theo:2}, the fine-tuning generalization error decreases as $N$ increases. Intuitively, transfer learning exhibits an economy of scale in the number of tasks: as the task portfolio grows, the model can pool more cross-task information, improving adaptation and reducing finetuning error. Because the domain gap is fixed for a given pretraining distribution, sufficiently informative finetuning data will eventually drive the finetuning error below the domain gap, at which point the overall estimation error is primarily governed by finetuning. Consequently, even when pretraining is poorly aligned with the target environment (i.e., domain knowledge is limited), the proposed method becomes increasingly effective as more real data accrue and can still deliver strong performance guarantees by learning directly from operational observations. 
More broadly, because our approach is directly inspired by LLM training pipelines, the quantitative theory developed here, though tailored to a small-data optimization setting, offers a principled lens on pretrain–finetune error mechanisms and helps understand when and why such paradigms perform well, including in Transformer-based LLMs.

\subsection{Pretraining, Finetuning, and Approximation Error Analysis}\label{sec:theo:2}

We begin by analyzing the domain gap, a component that is unique to the pretrain–finetune pipeline. The next theorem quantifies this gap in terms of the distributional mismatch between the true marginal $f_G$ and the pretraining marginal $f_{G^{\mathrm{pre}}}$, which encodes the decision maker’s domain knowledge. We measure this mismatch using the Hellinger distance, $H^2(f_G,f_{G^\mathrm{pre}})=\frac{1}{2}\int_{\R^p}\bigl(\sqrt{f_G(y)}-\sqrt{f_{G^\mathrm{pre}}(y)} \bigr)^2 dy$, which satisfies $0\leq H(f_G,f_{G^{\mathrm{pre}}}) \le 1$ by definition.

\begin{theorem}[Domain Gap]\label{thm:domain_gap}
    Suppose Assumption~\ref{ass:tail} holds. 
    Then there exists a constant $C>0$ such that
\begin{equation}\label{eq:domain_bound}
\mathcal{E}_{\mathrm{domain}}
\le
C H^2 \bigl(f_G,f_{G^{\mathrm{pre}}}\bigr)
\Big[
1-\log \Big(
H (f_G,f_{G^{\mathrm{pre}}})
\Big)
\Big]^{\max\{3,p\}}.
\end{equation}
\end{theorem}

Theorem \ref{thm:domain_gap} formalizes the intuition that more accurate domain knowledge, reflected in a smaller discrepancy between $f_{G^\mathrm{pre}}$ and $f_G$, leads to a smaller domain gap. In the idealized case where the pretraining and target distributions coincide, the domain gap vanishes; by Proposition~\ref{theo:interaction}, finetuning becomes unnecessary, and the pretrained estimator is already oracle-optimal. Even when pretraining is imperfect, the resulting initialization can still provide a useful starting point and yield informative estimates, particularly in regimes where finetuning is constrained by limited effective real data. Importantly, misaligned domain knowledge does not preclude strong performance: as task-specific data accumulate, the finetuning generalization error decreases, and the overall estimator improves accordingly. This perspective also underscores a practical limitation of a single universal pretrained model for all optimization domains: without adequate alignment, pretraining alone cannot guarantee optimality across heterogeneous settings. Nevertheless, pretraining remains valuable because it enables the decision maker to leverage large, domain-informed synthetic or historical corpora and substantial computational resources to build representations that meaningfully enhance performance in the small-data, large-scale regime.


Building on the decomposition in Proposition \ref{theo:interaction}, we next study the finetuning generalization error ${\mathcal{E}}_{\mathrm{gene}}$ for the pretrained Transformer estimator $\hat{\mathcal{T}}_{\textsf{TF}}$. To quantify how finetuning performance depends on the key ingredients of our pipeline, including the LoRA settings, the randomness in the real dataset, and the structure of the label-free finetuning objective (see Section \ref{subsubsec:finetuning} for details), we adopt an information-theoretic perspective inspired by \cite{xu2017information} and \cite{russo2019much}. Our results are novel in that they provide quantitative generalization guarantees for Transformer-based fine-tuning, which is the core architectural component of LLMs; to the best of our knowledge, this is the first comprehensive treatment along these lines.
Generally speaking, the information-theoretic approach relates generalization error to the mutual information between the model's inputs and outputs. When this mutual information is small, the model's outputs depend less on the idiosyncratic features of a particular dataset realization, thereby implying stronger out-of-sample generalization.

To establish our fine-tuning generalization bound, we require the following assumption (Assumption \ref{ass:subgaussian2}). In particular, a sub-Gaussian tail condition for the marginal $f_G$ follows from Assumption~\ref{ass:tail}; however, the resulting sub-Gaussian parameter can be cumbersome to track explicitly, so we state it directly for convenience. We make a similar simplification by assuming that the coordinates of $\bd\in\mathbb{R}^p$ are independent. This independence condition is not essential: the analysis readily extends to dependent coordinates with only minor modifications; see the online appendix.

\begin{assumption}\label{ass:subgaussian2}
    The marginal distribution $f_G$ is $\sigma^2_{d}$-sub-Gaussian. In addition, the coordinates of $\bd\in\mathbb{R}^p$ are independent of each other.
\end{assumption} 

\begin{theorem}[Finetuning Generalization Error]\label{theo_gene_bound}
 Suppose Assumption~\ref{ass:subgaussian2} holds. Consider a finetuning procedure in which a collection of weight matrices 
$\{W^{(\ell)}\in\mathbb{R}^{d^{(\ell)}_{\mathrm{out}}\times d^{(\ell)}_{\mathrm{in}}}\}_{\ell\in\mathcal{S}}$ is updated via the LoRA scheme introduced in Section \ref{sec:method}, where $\mathcal{S}$ denotes the set of layer indices selected for finetuning. Let $\{r_\ell\}_{\ell\in\mathcal{S}}$ be the corresponding retained ranks (cf. \eqref{eq:lora_1}), and define the total number of finetuned parameters as
$d_{\mathrm{finetuned}}
:=\sum_{\ell\in\mathcal{S}} r_\ell\big(d^{(\ell)}_{\mathrm{in}}+d^{(\ell)}_{\mathrm{out}}\big)$.
Then, there exists an absolute constant $C>0$ such that the finetuning generalization error satisfies  
   \begin{equation}\label{eq:Egene}
       {\mathcal{E}}_\mathrm{gene}
  \le
  C L_{\textsf{TF}}\Big(
\sigma^2+\sigma_d^2 [L_{\textsf{TF}}+e^{C/(L^4_{\textsf{TF}}\sigma^4_d)}]
\Big)\min\left\{
  \sqrt{\frac{ d_{\mathrm{finetuned}}}{N} },
  \frac{ d_{\mathrm{finetuned}}+1}{N}
  \right\},
   \end{equation}
   where the Transformer mapping is Lipschitz continuous with respect to the input, with Lipschitz constant denoted by $L_{\textsf{TF}}$ {(an explicit expression is provided in the online appendix.}
\end{theorem}

The proof is provided in the online appendix. Theorem \ref{theo_gene_bound} advances the existing literature by explicitly characterizing how key design choices in LoRA finetuning shape the generalization behavior of Transformer models. In particular, the retained low ranks $\{r_\ell\}_{\ell\in\mathcal{S}}$ and the selection of finetuning layers $\mathcal{S}$ serve as central hyperparameters: reducing either the ranks or the size of $\mathcal{S}$ decreases the number of trainable parameters and, consequently, tightens the generalization bound. 
At the same time, if the retained rank is chosen too small or the finetuning set $\mathcal{S}$ is overly restricted, the LoRA adaptation space may fail to contain descent directions that deliver nontrivial improvements, limiting the effectiveness of finetuning.

A further technical contribution underlying Theorem~\ref{theo_gene_bound} concerns the Lipschitz continuity of the Transformer mapping. Standard Transformer architectures are not Lipschitz by default, due in part to the instability of self-attention blocks and the use of normalization layers \citep{kim2021lipschitz}. Yet Lipschitz continuity plays a critical role in both training stability \citep{qi2023lipsformer} and theoretical analysis. To address this, we introduce two strategic architectural modifications: a Radical Clipping layer placed before the self-attention block and the replacement of standard normalization with CenterNorm; see Section \ref{subsec:transformer_architecture} and the online appendix for details. 
Importantly, Theorem~\ref{theo_gene_bound} shows that the finetuning generalization error essentially decays at a rate no slower than $\mathcal{O}(1/\sqrt{N})$. Thus, even in the small-data, large-scale regime, increasing the number of task instances $N$ yields more accurate estimates, bringing the fine-tuned model closer to the best-in-class solution within the Transformer hypothesis space; see Figure \ref{fig:overview}. This result highlights a key advantage of our approach: although each task instance may provide only limited information, the method effectively aggregates shared structure across a large collection of instances to improve individual decision-making.

We finally analyze the remaining component in Proposition \ref{theo:interaction}, namely the approximation error $\mathcal{E}_{\mathrm{approx}}$. Unlike canonical AI models that use only fully connected feedforward networks, Transformers incorporate structural modules, most notably the attention mechanism (see Section~\ref{subsec:transformer_architecture}), which are essential to their empirical success. These design features also make approximation analysis substantially more delicate, and existing results for standard DNNs (e.g., \citealt{han2023deep}) do not directly apply. Recall that our object of interest is the oracle mapping $\mathcal{T}^\ast$ that takes an input sequence of observations and returns an output sequence of parameter estimates, for arbitrary sequence length $\tau\in\mathbb{N}^+$. To study its approximability, we decompose this sequence-to-sequence mapping into interpretable segments that can be approximated separately. The following lemma motivates this decomposition and, importantly, mirrors the Transformer's modular structure.

\begin{lemma}[Kolmogorov-Arnold Representation Theorem \citealt{kolmogorov1957representations}]
\label{thm:rep-noPE} On a compact $\Omega\subset\mathbb{R}^p$, let 
$\mathcal{T}^\ast:\Omega^\tau\to(\mathbb{R}^p)^\tau$ map the input sequence
$\bm d=(\bd(1),\dots,\bd(\tau))$ to $\mathcal{T}^\ast(\bm d)=(\mathcal{T}^\ast_1(\bm d),\cdots,\mathcal{T}^\ast_\tau(\bm d))$, and denote the $k$-th coordinate of $\mathcal{T}^\ast_i$ as 
$\mathcal{T}^{\ast(k)}_i$ for $k=1,\ldots, p$.
  Then, for each $k\in\{1,\cdots,p\}$, there exist continuous functions
$ F^\ast\in C([0,1]^{2\tau p+1},\mathbb{R})$, $f^\ast\in C(\Omega,[0,1]^{2\tau p+1}),$ and $ \rho^\ast\in C(\Omega\times \Omega,\mathbb{R})$ (depending on the chosen $k$, with the dependence suppressed for notational simplicity)
such that, for every $\bm d\in \Omega^\tau$ and each $i\in\{1,\cdots,\tau\}$,
\begin{equation}
\label{eq:rep-noPE}
\mathcal{T}^ {\ast(k)}_i(d) = 
 F^\ast\Bigg(\sum_{j=1}^{\tau}
\mathrm{softmax} \big[ \rho^\ast\big(\bd(i),\bd(\cdot)\big)\big](j)  f^\ast \big(\bd(j)\big)\Bigg).
\end{equation}
\end{lemma}

We now explain how the representation in~\eqref{eq:rep-noPE} can be interpreted through the lens of the Transformer architecture introduced in Section~\ref{subsec:transformer_architecture}. For each entry $i=1,\cdots,\tau$, the data point $\bd(i)$ plays the role of the \emph{query}, and each $\bd(j)$ serves as a \emph{key} against which the query is compared. The resulting attention weight assigned to token $j$ is given by $\mathrm{softmax}\left[\rho^\ast \big(\bd(i),\bd(\cdot)\big)\right](j)$ in \eqref{eq:rep-noPE}. The corresponding \emph{value} is $f^\ast(\bd(j))$, and the weighted aggregation over $j$ forms an attention-style representation for position $i$.
Finally, $F^\ast$ acts as an output head that maps this representation to the $k$-th coordinate of the estimate. Lemma~\ref{thm:rep-noPE} therefore motivates a modular approximation strategy: approximate $\rho^\ast$, $f^\ast$, and $F^\ast$ separately, in a manner that mirrors the Transformer’s query–key–value structure. That said, the lemma is not directly applicable to our setting because it is stated on a compact domain, whereas our Transformer mapping is defined on an unbounded support. To proceed, we introduce two auxiliary definitions that characterize the function class to which the target mapping belongs.

\begin{definition}[Spectral Complexity]
\label{def:spectral-complexity}
For any square integrable function $\rho^*\in C(\Omega\times \Omega,\mathbb{R})$, its 
proper orthogonal decomposition (see the precise definition in \citealt{berkooz1993proper}) admits the expansion $ \rho^\ast(u,v)=\sum_{i\ge 1}\lambda_i \varphi_i^\ast(u)\psi_i^\ast(v)$, where $\lambda_1\ge\lambda_2\ge\cdots\ge0$ are the coefficients and $\{\varphi_i^\ast\}$ and $\{\psi_i^\ast\}$ are orthonormal bases in $L^2(\Omega)$. Fix $\alpha>0$ and for any $\mathcal{T}^ \ast$ having representations of the form \eqref{eq:rep-noPE}, we define its spectral complexity by $C_\mathrm{S}(\mathcal{T}^ \ast,\alpha)
:=\inf_{ F^\ast,f^\ast, \rho^\ast}\inf \{c>0: \lambda_s( \rho^\ast)\le cs^{-\alpha}, \forall s\ge 1\},$
where the outer infimum ranges over all admissible triples $( F^\ast,f^\ast, \rho^\ast)$ that represent $\mathcal{T}^ \ast$ in \eqref{eq:rep-noPE}.
\end{definition}

\begin{definition}\label{def:holder}
For $\beta, B > 0$ and a domain $\Omega \subset \mathbb{R}^p$, the Hölder function 
class $\mathcal{H}^\beta(\Omega, B)$ is defined as
\[
\mathcal{H}^\beta(\Omega, B)
:=
\Bigl\{
f : \Omega \to \mathbb{R}\ \Big|\ 
\max_{\|\bgamma\|_1 \le \lfloor \beta \rfloor}\|\partial^{\bgamma} f\|_\infty \le B,
\ 
\max_{\|\bgamma\|_1 = \lfloor \beta \rfloor}
\sup_{x \neq y}
\frac{|\partial^\gamma f(x) - \partial^\gamma f(y)|}{\|x - y\|^{\beta - \lfloor \beta \rfloor}}
\le B
\Bigr\},
\]
where
 $\bgamma = (\gamma_1,\ldots,\gamma_p)^\top$ denotes a multi-index of nonnegative integers with 
$\|\bgamma\|_1 := \sum_{i=1}^p \gamma_i$, and 
$\partial^{\bgamma} := \partial_1^{\gamma_1}\cdots \partial_p^{\gamma_p}$ denotes the corresponding mixed partial derivative.
\end{definition}

\begin{assumption}\label{ass:cs}
    (i) For some $\alpha>1$, the target mapping $\mathcal{T}^ \ast$ has finite $\alpha$-spectral complexity, i.e., $ C_\mathrm{S}(\mathcal{T}^ \ast,\alpha)<\infty$;  (ii) In the representation \eqref{eq:rep-noPE} of $\mathcal{T}^ \ast$, $\rho^*$ is square integrable. Moreover, the functions \(F^\ast\) and \(f^\ast\) in \eqref{eq:rep-noPE}, as well as the basis functions \(\{\varphi_s^\ast,\psi_s^\ast\}_{s\ge 1}\) appearing in the proper orthogonal decomposition of \(\rho^\ast\) in Definition~\ref{def:spectral-complexity}, belong to the Hölder function class $\mathcal{H}^\beta(\Omega,B)$ on their respective domains for some \(\beta,B>0\) (with \(\Omega\) allowed to vary across functions). The $F^\ast$ is Lipschitz continuous on its domain.
\end{assumption}

Assumption~\ref{ass:cs} imposes standard regularity conditions on the target mapping $\mathcal{T}^ \ast$, in the sense that it specifies a broad smoothness property rather than enforcing a restrictive parametric form. In particular, for a wide class of regular mappings, one can identify constants (e.g., the spectral complexity $\alpha$) under which Assumption~\ref{ass:cs} holds, making it a mild requirement. At the same time, such assumptions are necessary because any approximation guarantee must depend on baseline properties of the target mapping in order to quantify how well it can be approximated and to control the resulting approximation error. Such regularity conditions are standard in the approximation analysis of state-of-the-art AI models (see, e.g., \citealt{trans_approximation}). Under Assumption~\ref{ass:cs}, we next establish an approximation error bound. We use the notation  $a \asymp  b$ to indicate equality up to universal constant factors, i.e., $ a \asymp b  $ if there exist constants $ 0 < c' \le c'' < \infty  $ 
such that  $ c' b \le a \le c'' b  $.


\begin{theorem}[Approximation Error]\label{theo:approx}
Suppose Assumptions \ref{ass:tail} and  \ref{ass:cs} hold. 
Consider a Transformer architecture as described in Section \ref{subsec:transformer_architecture} with the following components: (\rmnum{1}) an embedding block implemented as an MLP 
with width $\mathcal{W}_{\mathrm{emb}}
\asymp(\lfloor\beta\rfloor+1)^2 3^{p} 
p^{\lfloor\beta\rfloor+2} N_{\mathrm{emb}}{\lceil\log_2(8N_{\mathrm{emb}})\rceil}$ and depth $\mathcal{D}_{\mathrm{emb}}
\asymp(\lfloor\beta\rfloor+1)^2 M_{\mathrm{emb}}{\lceil\log_2(8M_{\mathrm{emb}})\rceil} + p$, for any $M_{\mathrm{emb}},N_{\mathrm{emb}}\in\mathbb N_+$; (\rmnum{2}) attention heads with key and value dimensions $p_K$ and $p_V\ge 2 p+1$ (see Definition \ref{def:attention1}); and (\rmnum{3}) an output head implemented as an MLP with width $\mathcal{W}_{\mathrm{oh}}
\asymp(\lfloor\beta\rfloor+1)^2 3^{p_V} 
p_V^{\lfloor\beta\rfloor+1}p N_{\mathrm{oh}}{\lceil\log_2(8N_{\mathrm{oh}})\rceil}$ and depth $\mathcal{D}_{\mathrm{oh}}
\asymp(\lfloor\beta\rfloor+1)^2 M_{\mathrm{oh}}{\lceil\log_2(8M_{\mathrm{oh}})\rceil} + p_V$, for any $M_{\mathrm{oh}},N_{\mathrm{oh}}\in\mathbb N_+$, together with the radical clipping (RC) and CenterNorm layers in the pre-processing block (see the definitions in the online appendix).
Then the approximation error satisfies
\begin{equation}\label{eq:Eapprox}
\big[\mathcal{E}_{\mathrm{approx}}\big]^{1/2}
\le
C\Big[
p_V^{ \lfloor\beta\rfloor+(\beta\vee1)/2}
 (N_{\mathrm{oh}}M_{\mathrm{oh}})^{-\frac{2\beta}{p_V}}
+
p_K^{-(\alpha-1)}
+
(2p_K+p_V)^{\frac12} 
(N_{\mathrm{emb}}M_{\mathrm{emb}})^{-\frac{2\beta}{p}}
\Big],
\end{equation}
where $C$ is a constant depending only on the regularity of the target mapping (e.g., its spectral complexity). 
\end{theorem}

Theorem~\ref{theo:approx} shows that increasing model capacity through, for example, deeper or wider embedding and output-head networks, or larger attention dimensions $p_K$ and $p_V$, typically reduces the approximation error $\mathcal{E}_{\mathrm{approx}}$. This pattern is consistent with observations in other AI tools, such as DNNs. At the same time, obtaining an explicit, architecture-aware approximation bound for Transformers is technically nontrivial and, to our knowledge, has not previously been available in this form. Because Transformers constitute the core building block of modern LLMs, Theorem~\ref{theo:approx} also helps clarify why large models are empirically effective: greater capacity reduces approximation error, but exploiting that capacity requires sufficient training data. Without pretraining, enlarging the model entails a sharper trade-off: while approximation error decreases, the finetuning generalization error can deteriorate as model scale grows (see Theorem~\ref{theo_gene_bound}). Pretraining mitigates this tension by leveraging abundant synthetic or historical data to fit rich representations before exposure to scarce task-specific observations. Our framework inherits the same advantage: pretraining enables domain-knowledge-guided representation learning without being constrained by limited real data, and finetuning subsequently adapts the pretrained estimator to the operational environment while exploiting the transferable structure acquired during pretraining.


\subsection{From Estimation Error to Optimization Performance: Excess-Risk Bounds}\label{sec:final}

With the parameter estimates produced by our Transformer model, we can then solve the optimization tasks in \eqref{eq:formulation} simultaneously. In particular, cross-task information sharing is already encoded in these estimates, so each task reduces to a deterministic problem conditional on its estimated parameter. Specifically, for any $\btheta$, let $\bx(\btheta) := \arg\max_{\bx \in \mathcal{X}} v(\bx,\btheta)$ denote the optimal decision associated with parameter value $\btheta$. Under the ETO framework, we focus on the decision quality of the resulting data-driven solutions. For this purpose, we use an \emph{excess risk} criterion that benchmarks the induced policy against an oracle with access to the true prior. Given the observed data $\bbD$ and the pretrained-and-finetuned Transformer estimator $\hat{\mathcal{T}}_{\textsf{TF}}$, consider a fresh test instance $\bd$ generated according to \eqref{generating_process}. The resulting decision is $x(\hat{\mathcal{T}}_{\textsf{TF}}(\bd))$. In contrast, the oracle estimator is $\mathcal{T}^*$, inducing the oracle decision $x^\ast(\bd):=x(\mathcal{T}^*(\bd))$. We define the excess risk as the expected performance loss relative to the oracle:
\begin{equation}\label{eq:excess_opt}
    \mathcal{E}_{\mathrm{excess}}
    := \mathbb{E}_{\bbD} 
       \mathbb{E}_{\bd, \btheta}
       \big[v\big(x^*(\bd),\btheta\big)
       - v\big(x(\hat{\mathcal{T}}_{\textsf{TF}}(\bd)),\btheta\big)\big],
\end{equation}
where the inner expectation
$\mathbb{E}_{\btheta,\bd}$ is taken with respect to the hierarchical model in  \eqref{generating_process}, and the outer expectation $\mathbb{E}_{\bbD}$ averages over the randomness in $\hat{\mathcal{T}}_{\textsf{TF}}$, which depends on $\bbD$.

\begin{assumption}\label{ass:lip}
For any $\btheta $, the optimal decision $\bx(\btheta)$ uniquely exists and is $L_x$-Lipschitz continuous in $\btheta$; 
meanwhile, the value function $v(\bx,\btheta)$ is $L_v$-Lipschitz continuous in $\bx \in \mathcal{X}$.
\end{assumption}

Assumption \ref{ass:lip} is mild and is commonly satisfied in OM tasks; see two illustrative examples below. The existing ETO literature usually assumes that the objective is twice continuously differentiable in the model parameters, together with other regularity conditions (e.g., \citealt{OIC,ETOvsIEO}), which is technically comparable to Assumption \ref{ass:lip}.  


\begin{example}[Large-Scale Newsvendor]\label{example:newsvendor}
Consider a firm that manages inventories for a large collection of products indexed $i = 1, \ldots, N$. Let the decision vector be $\bx = (x_1, \cdots, x_N) \in \mathcal{X}$, where $x_i$ denotes the order quantity for product $i$. The random demand vector is $\bd=(d_1,\ldots,d_N)$, with each component distributed as $d_i\sim \mathcal{N}(\theta_i,\sigma^2)$.
The expected newsvendor loss for product $i$ is given by 
$v(x_i, \theta_i)
= \mathbb{E} 
[ b_i (d_i - x_i)_{+}
    + h_i (x_i - d_i)_{+}
],$
where $b_i$ and $h_i$ represent the unit shortage and holding costs, respectively. This formulation is similar to the setting in \cite{mukherjee2015efficient}. The optimal order quantity for each product admits a closed-form solution, $x_i(\theta_i)
= \theta_i
+ \sigma
  \Phi^{-1}(\frac{b_i}{b_i+h_i})$, where $\Phi^{-1}(\cdot)$ is the inverse cumulative distribution function of the standard normal distribution. It follows immediately that $x_i(\theta_i)$ is Lipschitz continuous in $\theta_i$.
\end{example}

\begin{example}[Large-Scale Pricing]
Consider a firm that sells a large range of products and must simultaneously determine pricing decisions.
For illustrative purposes, suppose the random demand for product $i$ follows $d_i \sim\mathcal{N}( \theta_i -\nu_i x_i , \sigma^2)$, 
where $\theta_i$ represents the market size and $\nu_i>0$ is a known price-sensitivity coefficient. The individual expected revenue  is then
$v(x_i, \theta_i) 
= \mathbb{E}
[ x_i d_i
]
= \big(x_i \theta_i - \nu_i x_i^2\big).$
Maximizing this objective yields the optimal price for each product, $x_i(\theta_i) = \theta_i/(2\nu_i)$, which is Lipschitz continuous in $\theta_i$.
\end{example}

We conclude by analyzing the excess risk in the following theorem. The key trade-off is between the finetuning generalization error in Theorem~\ref{theo_gene_bound}, which typically increases with model complexity, and the approximation error in Theorem~\ref{theo:approx}, which decreases as the model becomes richer. By selecting a network configuration that balances these two effects, we obtain the best achievable convergence rate as a function of the number of simultaneous tasks, $N$, thereby quantifying the efficiency of cross-task transfer. This small-data, large-scale perspective differs from the classical big-data regime, in which accuracy improves primarily by increasing the number of observations per task. The optimal model size also depends on which Transformer modules are adapted via LoRA. A recent highly influential empirical study by \cite{schulman2025lora} reports that applying LoRA to attention layers delivers little incremental benefit beyond finetuning the MLP components (i.e., the embedding layer and output head). Our experiments in the online appendix are consistent with this finding. Accordingly, the theorem below focuses on the setting in which LoRA finetuning is applied only to the embedding layer and output head, while the attention block remains frozen.

\begin{theorem}[Excess Risk]\label{theo:final}
Suppose that Assumptions~\ref{ass:lora}-\ref{ass:lip} hold, and that LoRA with rank $r$ is applied to all layers
in the embedding block and the output head.
 A rate-optimal choice of model size is given by
$$N_{\mathrm{emb}}M_{\mathrm{emb}}\asymp N^{\frac{p}{p+8\beta}}(p_K)^\frac{2p}{p+8\beta},\quad N_{\mathrm{oh}}M_{\mathrm{oh}}\asymp 
N^{\frac{p_V}{p_V+8\beta}},\quad p_K\asymp N^{\frac{2\beta}{(p_V+8\beta)(\alpha-1)}},$$
with $p_V=\max\{2 p+1,\frac{\alpha-1/2}{\alpha-1}p\}.$
Under this design, up to logarithmic factors, the excess risk satisfies
\begin{equation}
     \mathcal{E}_{\mathrm{excess}}\le CL_vL_x\bigg(\min\big\{ C_{\mathrm{gene}}\sqrt{{r}}N^{-\frac{4\beta}{p_V+8\beta}} 
 ,\mathcal{E}_{\mathrm{domain}}\big\} + {p_V^{ 2\lfloor\beta\rfloor+(\beta\vee1)}  N^{-\frac{4\beta}{p_V+8\beta}}}
 \bigg).
\end{equation}
 where $C_{\mathrm{gene}}=L_{\textsf{TF}}\Big(
\sigma^2+\sigma_d^2 [L_{\textsf{TF}}+e^{C/(L^4_{\textsf{TF}}\sigma^4_d)}]
\Big)$, and the constant $C>0$ depends on the complexity of the target mapping. The domain gap $\mathcal{E}_{\mathrm{domain}}$ is bounded as in~\eqref{eq:domain_bound}.
\end{theorem}

This result demonstrates that our approach effectively addresses the large-scale, small-data optimization setting and exhibits an economy-of-scale effect: performance improves as the number of tasks $N$ increases. Under the rate-optimal model configuration, it attains a convergence rate of $\mathcal{O}(N^{-\frac{4\beta}{p_V+8\beta}})$ convergence rate, with $p_V=\max\{2\tau p+1,\frac{\alpha}{\alpha-1}p\}$. The rate is faster for smoother targets (larger $\beta$) and for targets of lower spectral complexity (larger $\alpha$).


\section{Simulation Studies}\label{sec:numerical}

In this section, we conduct simulation experiments to examine the effectiveness of the proposed pretrain-finetune paradigm. Beyond documenting numerical performance, our primary objective is to verify the theoretical insights on the interaction between pretraining and finetuning developed in Section~\ref{sec:theory}. We focus on three implications. First, when real observations are scarce, pretraining is critical for enabling the effective training of large Transformer models, and the quality of the resulting initialization depends on the distributional proximity between the pretraining distribution and the true data-generating distribution, i.e., on the accuracy of the decision maker’s domain knowledge. Second, when this domain knowledge is well aligned with practice, the pretrained model alone may be sufficient, and additional fine-tuning offers little incremental value. Third, even under misspecified domain knowledge, the proposed framework can still address the small-data, large-scale regime by transferring information across many instances, thereby reducing excess risk as the number of tasks $N$ increases. Because our focus is the pretrain–finetune mechanism and given space constraints, we leave empirical evaluations on real-world datasets to future work.

For illustration, we consider the large-scale multi-product newsvendor setting in Example~\ref{example:newsvendor}. There are $N$ products, and for each product $i$ the decision maker observes a single demand realization $d_i \sim \mathcal{N}(\theta_i,1)$, where $\theta_i\in\mathbb{R}$. We set $b_i=h_i=2$. Based on these observations, the decision maker selects an order quantity for each product. In Section~\ref{sec6.1}, to make the domain gap easier to quantify, we first study a simple setting in which the oracle prior for $\theta_i$ follows a Gaussian mixture model, while the decision maker’s domain knowledge is represented by a misspecified Gaussian mixture with possibly shifted component locations. In Section~\ref{sec6.2}, we extend to richer data-generating processes for $\theta_i$, spanning both parametric families and flexible nonparametric specifications, to assess the robustness of the numerical patterns.

\paragraph{Transformer Model Architecture.} Our Transformer uses an embedding dimension of $p_{\mathrm{emb}} = 64$, and the multi-head self-attention
module uses $8$ attention heads. The embedding block contains two layers, and the output head comprises $24$ layers, yielding approximately $5\times10^4$ trainable parameters in total. 
All experiments were run on a workstation equipped with four NVIDIA GeForce RTX 4090 GPUs (24GB each).

\subsection{Pretrain-Finetune Interaction}\label{sec6.1}

In what follows, we first quantify how the pretrained model's performance varies with the accuracy of the decision maker’s domain knowledge. We then examine how finetuning on real observations improves pretrained models across different levels of domain-knowledge quality. Finally, we study the role of the problem scale $N$, which in our setting is the effective finetuning sample size.

\textit{Target and pretraining distributions.} We take the task parameters to follow a three-component Gaussian mixture, i.e., $\theta_i \sim G$, where $G= \tfrac{1}{3}\mathcal{N}(1,1) + \tfrac{1}{3}\mathcal{N}(3,1) 
+ \tfrac{1}{3}\mathcal{N}(4,1).$ The decision maker’s domain knowledge could be imperfect and posits that demand is concentrated around three locations $\tilde{\theta}_1,\tilde{\theta}_2,\tilde{\theta}_3$. Accordingly, the pretraining prior is specified as $G^{\mathrm{pre}}=\sum_{j=1}^{3} \tfrac{1}{3} \mathcal{N}(\tilde{\theta}_j,1)$. To emulate heterogeneous levels of domain-knowledge accuracy, we construct $15$ pretrained models by varying $(\tilde{\theta}_1,\tilde{\theta}_2,\tilde{\theta}_3)$, drawing each $\tilde{\theta}_j$ independently from $\mathrm{Unif}[0,5]$. Closed-form expressions for the Hellinger distance or KL divergence between Gaussian mixtures are generally unavailable. Nevertheless, as noted by \citet{goldberger2003efficient}, when mixture weights and component variances coincide, the discrepancy between two mixtures can be upper bounded by the $\ell_2$ distance between their component-mean vectors. We therefore use $\big\|(\tilde{\theta}_1,\tilde{\theta}_2,\tilde{\theta}_3) - (1,3,4)\big\|$ as a tractable surrogate for the distributional distance between the pretraining distribution and the target
distribution.

\textit{Pretraining and finetuning setup.}
In pretraining, each synthetic data sequence $\bbD^{\mathrm{pre}}_k:=(\bd_{k,1}, \dots, \bd_{k,512})$ has fixed length $512$, with $k$ indexing different sequences. We pretrain all models with Adam (batch size $32$) for $1000$ iterations. This yields approximately $512 \times 32 \times 1000 \approx 1.6 \times 10^7$ synthetic observations with orders of magnitude more than the real sample, thereby providing a stable and meaningful initialization for the Transformer training. We then finetune each pretrained model on $N=500$ real observations generated from the target distribution. For LoRA, we set the retained rank to $r=8$, freeze all attention-module parameters, and update only the embedding and output-head components, amounting to roughly $6\times 10^3$ trainable parameters. Finetuning is run for $10$ epochs. To motivate these design choices and offer practical guidance, we conduct extensive ablation studies in the online appendix. The results indicate that $r=8$, and often even smaller ranks, suffices for effective adaptation in our setting. Moreover, finetuning only the embedding and output-head modules delivers essentially the full performance gains, with no clear benefit from finetuning all components, consistent with \cite{schulman2025lora}.

\begin{figure}[htbp]
    \centering
    \includegraphics[width=0.66\linewidth]{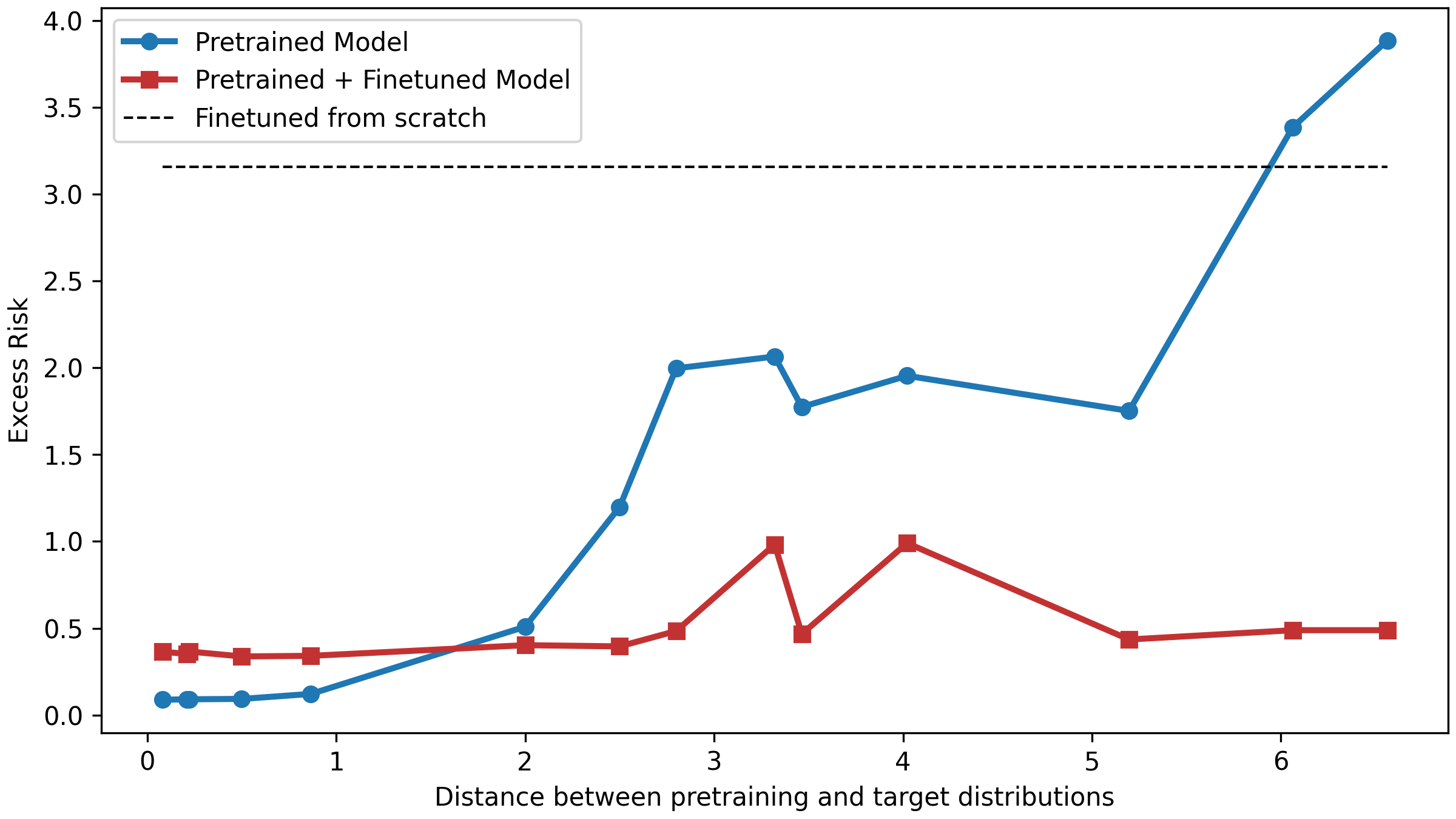}
\caption{Excess risk of pretrained-only, pretrained-finetuned, and finetuned-from-scratch models versus the distributional distance between the pretraining and target distributions.}
    \label{fig:exp_1}
\end{figure}

\textit{Effect of domain knowledge accuracy.}
Figure~\ref{fig:exp_1} reports the excess risk for the 15 pretrained models and their finetuned counterparts. We estimate excess risk by averaging realized performance over $500$ independent test instances drawn from the target distribution. The horizontal axis measures the distributional distance between the target and pretraining distributions; moving from left to right corresponds to less accurate domain knowledge. Here, ``Pretrained + Finetuned" refer to models that are pretrained on synthetic datasets of varying quality and then finetuned using the real observations, whereas “Pretrained” evaluates the same models immediately after pretraining, without any finetuning. We also include a ``Finetuned from scratch” baseline, which trains an unpretrained Transformer (initialized with PyTorch defaults) using only the real data. Since this baseline does not depend on the pretraining distribution, it appears as a horizontal line in Figure~\ref{fig:exp_1}.

Several interesting patterns emerge in Figure~\ref{fig:exp_1}. First, as predicted by Theorem~\ref{thm:domain_gap} and Proposition~\ref{theo:interaction}, the pretrained-only estimator degrades as the pretraining distribution moves farther from the target distribution, implying that weaker domain knowledge translates into higher excess risk. Second, once fine-tuned on real observations, all models achieve uniformly low excess risk, largely independent of pretraining quality. This insensitivity suggests that, after finetuning, the generalization component dominates the decomposition in Proposition~\ref{theo:interaction}, so residual variation in the domain gap plays a limited role in realized performance. Interestingly, when the distance between the pretraining and target distributions is below $2$, pretraining alone is often sufficient as the pretrained-only model performs comparably to, and even better than, the finetuned version. This regime is consistent with Proposition~\ref{theo:interaction}, indicating that sufficiently accurate domain knowledge can render additional adaptation unnecessary. Lastly, the ``finetuned-from-scratch'' baseline performs markedly worse than the pretrained-and-finetuned models. The gap underscores the practical importance of pretraining even when domain knowledge is imperfect: the available real data are too limited to reliably train a large Transformer from random initialization, whereas pretraining supplies a well-structured starting point that makes subsequent finetuning both feasible and effective.

\begin{figure}[htbp]
    \centering
    \includegraphics[width=0.86\linewidth]{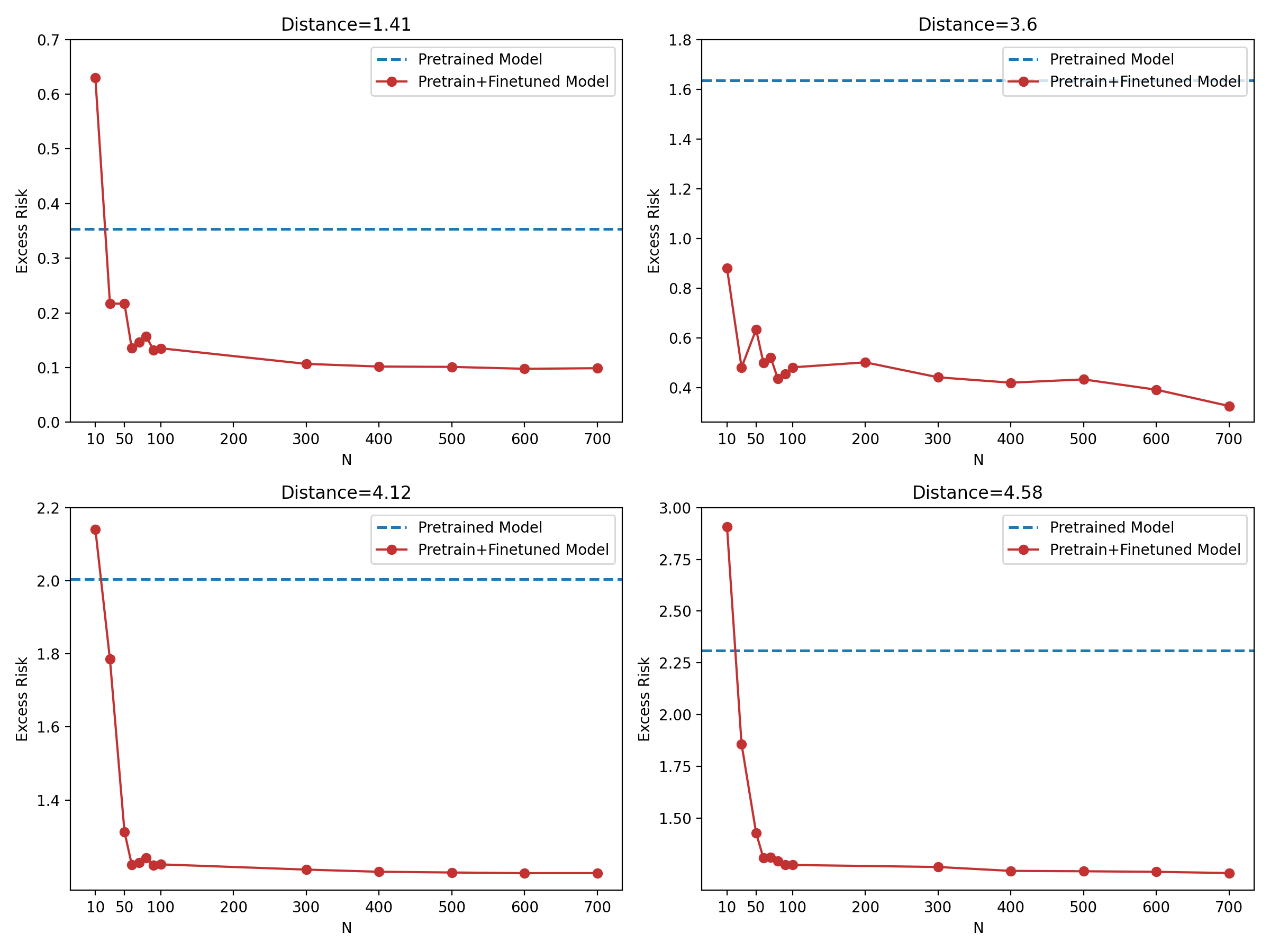}
   \caption{Excess risk versus finetuning data size $N$ under four pretraining configurations.}
    \label{fig:effectofN}
\end{figure}

\textit{Effect of the number of problem instances $N$}.
We next examine how performance varies with the finetuning sample size $N$. We select four pretrained models whose pretraining distributions have different distances to the target distribution (specifically, $1.41$, $3.60$, $4.12$, and $4.58$). For each model, we vary the finetuning sample size from $N=10$ to $N=700$ and finetune accordingly. Figure~\ref{fig:effectofN} reports the excess risk of the pretrained-only models and their pretrained–finetuned counterparts. Across all four configurations, the pretrained-only models exhibit consistently higher excess risk, whereas the pretrained–finetuned models improve steadily and their excess risk declines as $N$ increases. This pattern aligns with Theorem~\ref{theo:final} and demonstrates the value of the proposed approach in large-scale regimes, where a larger pool of instances strengthens the transfer-learning effect. Meanwhile, Theorem~\ref{theo:approx} implies an irreducible approximation component determined by model capacity; accordingly, in our experiments, the excess risk does not fully disappear even at the largest $N$, plausibly reflecting residual approximation error from the finite Transformer architecture, which may not represent the oracle estimator exactly.

\subsection{Robustness to More Complex Target Distributions}\label{sec6.2}

We further assess the robustness of the preceding findings in more challenging environments.  
To better reflect practical demand heterogeneity, we consider a range of data-generating processes for $\theta_i$, including both simple parametric families and flexible nonparametric specifications. In particular, we include:
\begin{itemize}
    \item {Exponential distribution:}
    $\theta_i$ follows an exponential distribution with mean $5$, which exhibits a heavier right tail than the Gaussian-mixture case.
\item {Dirichlet process (DP) with Uniform base:}
We consider a DP prior with concentration parameter $\alpha_{\mathrm{DP}} = 1$ 
and base distribution $\mathrm{Unif}[0,5]$. Equivalently, 
$\theta_1,\ldots,\theta_N$ are generated according to:
$$\theta_j =
\begin{cases}
\theta_i, & \text{with probability } \dfrac{j-1}{\alpha_{\mathrm{DP}} + j - 1}, \quad i \in \{1,\ldots,j-1\},\\[3mm]
x \sim \mathrm{Unif}[0,5], & \text{with probability } \dfrac{\alpha_{\mathrm{DP}}}{\alpha_{\mathrm{DP}} + j - 1}.
\end{cases}$$
This construction induces flexible distributions for $\theta_i$ that can depart substantially from standard parametric families.

    \item {Dirichlet process with Gaussian base:}
    A DP prior with $\alpha_{\mathrm{DP}} = 1$ and a Gaussian base distribution $\mathcal{N}(2.5,1)$. 

    \item {Neural distribution:}
    Following \citet{EB+TF}, we also consider a highly flexible, nonparametric data-generating mechanism constructed from randomly initialized neural networks. Specifically, we sample four two-layer networks, each mapping $\mathbb{R}^4$ to $\mathbb{R}$, with weights initialized using PyTorch’s default scheme. For each hidden layer, the activation function is drawn uniformly from \texttt{GELU}, \texttt{ReLU}, \texttt{SELU}, \texttt{CELU}, \texttt{SiLU}, and \texttt{TanhShrink}, and the final layer uses a sigmoid activation. To generate $\theta_i$, we draw an input vector uniformly from $[0,1]^4$, select one of the four networks uniformly at random, and take its scalar output as the realization of $\theta_i$.
\end{itemize}

\begin{figure}[htbp]
    \centering
    \includegraphics[width=0.86\linewidth]{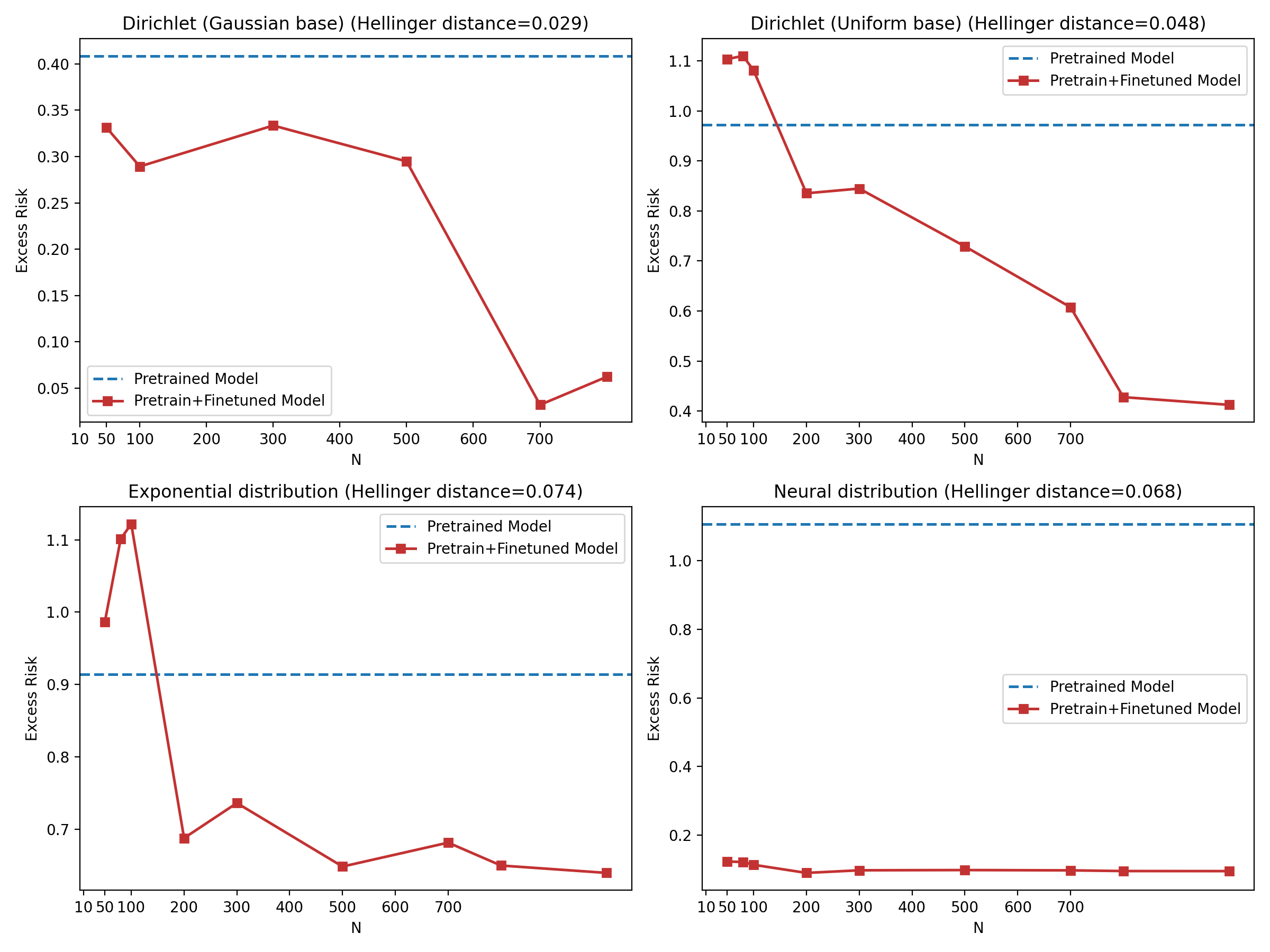}
   \caption{Excess risk of the pretrained-only and pretrained-finetuned models versus the finetuning data size $N$ across different target distributions.}
\label{fig:exp2}
\end{figure}

Across all four target distributions, we endow the decision maker with domain knowledge of the same form as in Section~\ref{sec6.1}, namely a three-component Gaussian-mixture pretraining distribution, $\tfrac{1}{3}\mathcal{N}(1,1) + \tfrac{1}{3}\mathcal{N}(3,1) + \tfrac{1}{3}\mathcal{N}(4,1)$. We pretrain the Transformer on this distribution using the same pretraining protocol as in Section~\ref{sec6.1}, and then finetune the resulting model on task-specific observations drawn from the target environment. We again vary the number of instances from $N=50$ to $N=1000$ and track the resulting excess risk. Figure~\ref{fig:exp2} reports the performance across target distributions; for reference, we also report Monte Carlo estimates of the Hellinger distance between each target distribution and the pretraining distribution.

Consistent with Theorem~\ref{theo:final} and the evidence in Section~\ref{sec6.1}, the proposed approach performs robustly across all distributional settings: the excess risk of the pretrained–finetuned estimator declines as the number of instances $N$ increases, demonstrating effective learning in the large-scale, small-data regime. Moreover, for the exponential and Dirichlet-process targets with a Uniform base measure, the pretrained-only estimator can outperform its fine-tuned counterpart when $N$ is extremely small. This phenomenon suggests that, with only a handful of task-specific observations, the structure transferred via pretraining may be more informative than adapting the model using highly scarce data. As $N$ grows, however, finetuning rapidly becomes beneficial and the pretrained–finetuned model overtakes the pretrained-only baseline.

The advantage of pretraining followed by finetuning is most pronounced under the Neural distribution, which embodies a highly complex, nonparametric data-generating process. In this setting, the pretrained–finetuned estimator improves rapidly with only a modest amount of real data and substantially outperforms the pretrained-only baseline. The reason is that the three-component Gaussian-mixture prior used for pretraining, intended to encode the decision maker’s domain knowledge, does not capture the salient structure of the Neural target distribution. Consequently, strong performance must be driven primarily by information learned from task-specific observations through finetuning. This finding highlights the value of the proposed pretrain–finetune framework when domain knowledge is misspecified.

\section{Conclusion}\label{sec:conclusion}

In this paper, we propose to solve the small-data, large-scale stochastic optimization problems using an LLM-inspired
pretrain-then-finetune approach. We implement the idea in a Transformer model designed for this purpose, which is naturally compatible with the pretrain-finetune pipeline and leverages the attention mechanism to deliver state-of-the-art representational capacity. On the theory side, we develop, to the best of our knowledge, the first comprehensive error analysis for Transformer learning in this decision-centric setting. The results quantify how performance depends on (i) the \emph{domain gap} induced by mismatch between the pretraining distribution and the operational environment, (ii) the \emph{finetuning} sample size, which in our formulation scales with the number of tasks and captures an economics-of-scale effect in transfer learning, and (iii) architectural choices that govern approximation capacity. The analysis also clarifies the interaction between pretraining and finetuning: when effective real data are scarce, domain-knowledge-guided pretraining can be the primary driver of accuracy, whereas with sufficiently many task instances, finetuning progressively dominates and corrects pretraining bias. Beyond their implications for small-data optimization, these insights contribute to a principled understanding of why pretrain-finetune pipelines can be effective more broadly. Looking ahead, we view the proposed framework as a step toward small, specialized models for operational decision problems that can leverage both domain knowledge and limited proprietary data. Promising directions for future research include evaluating the approach on large-scale field datasets and developing principled, operationally grounded procedures for constructing domain-informed pretraining distributions in practice.


%
%


\bibliographystyle{informs2014} 
\bibliography{ref} 

\end{document}